\documentclass[Afour,sageh,times]{ieeeconf}

\pdfoutput=1

\usepackage{mdwlist}

\usepackage{graphics} % for pdf, bitmapped graphics files
\usepackage{epsfig} % for postscript graphics files
\usepackage{mathptmx} % assumes new font selection scheme installed
\usepackage{amsmath} % assumes amsmath package installed
\usepackage[utf8]{inputenc}
\usepackage{graphicx}
\usepackage{tikz}
\usepackage{tabularx}
\usepackage{caption}
\usepackage{amsfonts}
\usepackage{amssymb}
\usepackage{bm}

\usepackage[noadjust]{cite}
\usepackage{epstopdf}
\usepackage{url}

\usepackage[colorlinks,bookmarksopen,bookmarksnumbered,citecolor=red,urlcolor=red]{hyperref}

\newcommand\BibTeX{{\rmfamily B\kern-.05em \textsc{i\kern-.025em b}\kern-.08em
T\kern-.1667em\lower.7ex\hbox{E}\kern-.125emX}}

\begin{document}

\title{A new time-projecting controller based on 3LP model to recover intermittent pushes}

\author{Salman Faraji$^{*}$ and Auke J. Ijspeert$^{*}$ \thanks{$^{*}$Biorobotics Laboratory, Ecole Polytechnique F\'ed\'erale de Lausanne (EPFL),  Lausanne, Switzerland}}

\maketitle

\begin{abstract}
In this paper, we present a new walking controller based on 3LP model. Taking advantage of linear equations and closed-form solutions of 3LP, the proposed controller can project the state of the robot at any time during the phase back to a certain event for which, a discrete LQR controller is designed. After the projection, a proper control policy is generated by the expert discrete controller and used online. This projecting architecture reacts to disturbances with minimal delay and compared to discrete controllers, it provides superior performance in recovering intermittent external pushes. Further analysis of closed-loop eigenvalues and disturbance rejection shows that the proposed time-projecting controller has strong stabilization properties. Controllable regions also show that the projecting architecture covers most of the maximal controllable set of states. It is computationally much faster than model predictive controllers, but still optimal.
\end{abstract}

\section{Introduction}

Performing bipedal locomotion on humanoid robots is a challenging task regarding many different aspects. On one hand, the hardware should be powerful enough to handle the weight and fast motions of the swing legs. On the other hand, controllers require precise perception and actuation capabilities for better stabilization and predictability of the system. Power and precision are two different and sometimes conflicting requirements. From the perspective of geometry also, complex chains in each limb of the robot make the control problem more sophisticated. For slow walking speeds, footstep locations and proper ankle torques become very important for stabilization \cite{faraji2014robust}. In faster speeds however, the system has an intrinsic self-stabilizing properties \cite{rummel2010stable}. Proper tracking of desired joint angles or Cartesian trajectories therefore seem to be critical. In this regard, hierarchical model-based control approaches can handle complexities in different levels and use models to capture main dynamics of the actuators, limbs or the full body. 

\subsection{Template models}

In higher control levels, due to complexity of the full model, it is conventional to use lower-dimensional template models which provide abstract dynamics. These models can speed up calculation of footstep plans, Center of Mass (CoM) and Center of Pressure (CoP) trajectories. Inverted Pendulum (IP) \cite{kuo2005energetic} is probably the simplest model, concentrating the whole mass of the robot in a point and modeling the legs with massless inverted pendulums. In this model, swing and torso dynamics are absent and therefore, the timing or the final attack angle is imposed by the controller, hoping that lower level controllers can track this motion. There are more complex versions of inverted pendulum with masses in the legs \cite{byl2008approximate}, torso \cite{westervelt2007feedback} and knee for swing leg \cite{asano2004novel}. In all advanced versions of IP, it is possible to obtain a natural gait, but only through a constrained optimization and numerical integration due to non-linearity. It is often popular to linearize these systems around a pre-calculated gait and use a discrete linear model for control \cite{rummel2010stable}. One should therefore create a library of optimal primitives to handle different gait conditions \cite{kelly2015non, manchester2014real, gregg2012control}.

The linear version of inverted pendulum (LIP) however provides analytical solutions and is widely used in slow-walking locomotions \cite{feng20133d, faraji2014robust, herdt2010walking}. In this model, similar to IP, the next footstep location and timing should be imposed as the original model does not include swing dynamics. However, one can introduce double support phases and handle the weight transfer smoothly, unlike impacts in the original IP. A disadvantage however is the constant CoM height and thus, crouched knees.

\subsection{Advantages of 3LP}

The 3LP model introduced in \cite{faraji20163LP} provides the same linear properties which are favorable for control. But thanks to inclusion of swing dynamics in 3LP, one can calculate periodic gaits analytically. Torso-balancing hip torques are also included in 3LP, hence one can expect more natural CoM trajectories compared to the LIP model. These trajectories are therefore easier to track for lower level controllers in the hierarchy. Since 3LP can produce abstract gaits itself, we do not need to impose attack angles or footstep locations anymore. Note that these gaits are expressed in abstract level, only describing feet and CoM trajectories. Handling internal joint coordinates and complexities of the full model is left for lower-level controllers, like inverse dynamics \cite{faraji2014robust}. Finally, unlike discretization of non-linear models which provide a discrete step-to-step control framework, thanks to the linearity and closed-form solutions of 3LP, one can setup continuous control for any time during the phase. In this article, we are going to exploit this property and introduce a better time-step control framework which is more responsive to external perturbations. 

\subsection{Limitations of discrete control}

Bipedal walking naturally consists of hybrid phases, i.e. single and double support where the continuous model of the system differs in each phase. Hybrid switching happens either in heel-strike (touch down) or push off (toe off) events. Original IP models \cite{kuo2005energetic} have impacts at these events while linear versions can handle the transition smoothly. In advanced IP-based models due to nonlinearity, one needs to use numerical integration to calculate a linear map associated to a certain discrete event like heel-strike or maximum apex height. Although geometries and momentum laws can help finding hybrid transition conditions for IP \cite{zaytsev2015two, kelly2015non, byl2008approximate}, controlling the system during the continuous phases is rarely addressed in the literature. A possible reason could be the need to predict state evolution until the end of the phase which requires numerical integration. In LIP and 3LP however, closed form solutions can speed-up this process. But one needs to find a way to properly handle the interplay between continuous and discrete variables. For example, hip and ankle torques can modulate the CoM speed during single support phase, but they also influence the next foot-step location which remains fixed over the whole next step and serves as a discrete variable.

\begin{figure}[]
        \centering
        \includegraphics[page=3,trim = 0mm 7mm 0mm 0mm, clip, width=0.5\textwidth]{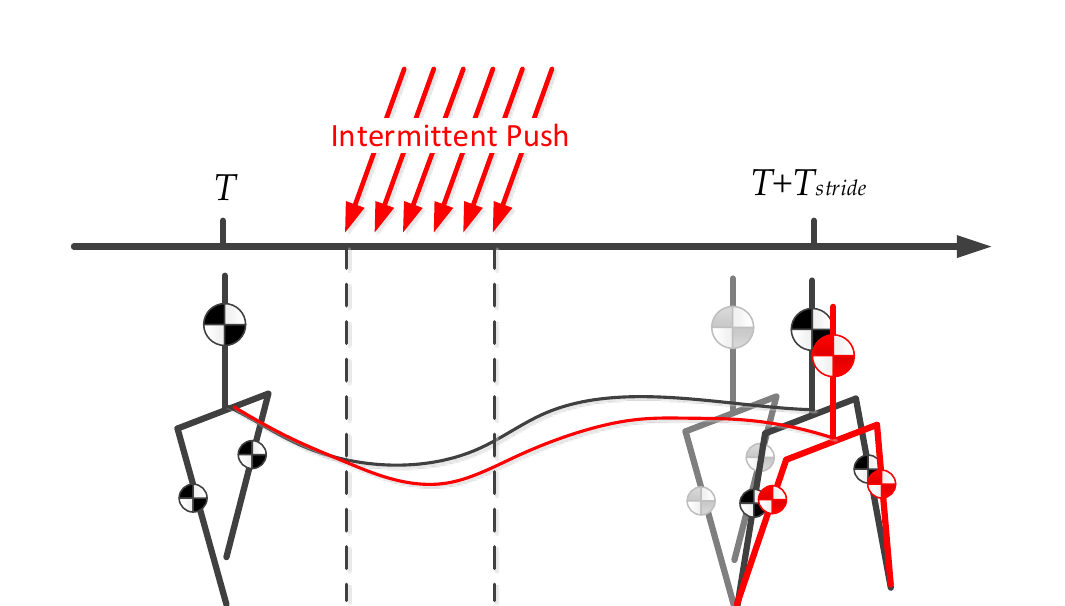}
        \caption{Demonstration of an intermittent push that appears shortly during a continuous phase and influences the system. The normal and disturbed trajectories are shown in black and red respectively. A delayed reaction to such disturbance might produce a large overshoot in the next footsteps.} 
        \label{fig::intermittent_push}
\end{figure}

\subsection{Continuous control}

In the present work, we focus on the availability of closed-form solutions in 3LP to setup a continuous and online control paradigm. The aim is to react to intermittent disturbances as soon as they take place, as opposed to discrete controllers which have to wait until the next step. Intermittent disturbances can have different magnitude, duration and timing. Normally, earlier pushes can disturb the system more severely due to exponential nature of falling dynamics (Figure.\ref{fig::intermittent_push}). Traditional Poincar\'e based methods \cite{poincare} which linearize the system around a pre-optimized gait cannot capture continuous effects. In other words, all disturbances happening between two discrete events are accumulated and observed only at the next event. To setup a continuous control however, there should be a measure to evaluate the error in the middle of continuous phases. In addition, one also needs to know the effect of available continuous inputs (like hip/ankle torques) on the final discrete variables (like footstep location). Similarly, online adjustment of attack angles has been used in hopping and running algorithms with simple \cite{raibert1984experiments} or more complex \cite{faraji2013compliant} models. Such online paradigm let the robot react to unforeseen disturbances as fast as possible to avoid taking large steps which can happen if the reaction is delayed. However in these methods, the design of control law is either by systematic search or intuitive tuning, using numerical integrations. Unlike using Poincar\'e maps, there is no systematic way to take the best inter-phase optimal reaction online, for any gait speed and disturbance observed. Although offline optimizations and primitive libraries might work, we want to propose a generic control architecture that provides the best online reaction. 

\subsection{Time-projecting controller}

Closed form solutions of 3LP let us design a single discrete LQR (DLQR) \cite{ogata1995discrete} controller which is optimal for all kinds of gaits that 3LP can produce. Discrete architectures are proposed a lot in literature \cite{kelly2015non, byl2008approximate, manchester2014real, sharbafi2012controllers}. In these works, a certain event is considered to decide the new angle of attack or push-off force. Our DLQR has a similar functionality and serves as a core stabilizing expert in our proposed architecture, despite being designed for one specific moment, i.e. foot touch-down. For any time in the middle of continuous phases then, we project or map the currently observed error back in time to the touch-down event, where DLQR controller knows best how to handle it. We take the output of DLQR then and apply it to the system in the middle of the phase. Such online policy makes sure that the future evolution of the system given the calculated input will be optimal, if seen in a bigger time-span over multiple future steps. We also take advantage of a simple disturbance observer to decouple internal dynamics from external pushes. With certain assumptions, our proposed controller captures the extra energy injected by external pushes and stabilizes the system. Because of unstable falling dynamics, a push of certain magnitude and duration might have different effects if applied early or late during single support. Our continuous time-projecting controller (CTPC) is therefore expected to handle such sensitivity to timing, i.e. handling continuous disturbances.

\subsection{Comparison to MPC}

An alternative to CTPC could be setting up a model predictive controller (MPC) similar to our previous work \cite{faraji2014robust}. Such MPC controller can consider the remaining time of the current continuous phase as well as few next steps to stabilize the system and handle intermittent internal pushes. Additionally, using 3LP, such MPC controller can incorporate inequality constraints on input torques, center of pressure, friction cones and footstep length to ensure feasibility of the plan. Motivated by observations and mathematical analysis that postulate sufficiency of considering maximum two steps for future planning \cite{zaytsev2015two}, the alternative MPC controller might not be computationally very expensive too. In this work however, compared to \cite{faraji2014robust,feng20133d, kuindersma2014efficiently} we propose a much simpler and faster framework. It does not consider inequality constraints, but thanks to analysis of controllable regions \cite{zaytsev2015two}, we can provide a simple criteria that indicates whether the robot should switch to a more complex controller or even emergency cases. Further analysis of controllable regions indicates that extreme conditions rarely happen in slow frequencies and small stride lengths. Therefore, CTPC is enough most of the time and if not, all other controllers including MPC can hardly stabilize any better. We would also like to remark that 3LP already describes the pelvis width and thus avoids internal collision in natural walking. In extreme cases where humans cross over laterally, a MPC controller with non-convex constraints or advanced collision avoidance algorithm might be needed. 

This article starts with the derivation of discrete-error dynamic equations over touch-down event. After calculation and analysis of different DLQR controllers in the next section, we go forward and introduce the core idea of time-projection in our continuous-time controller. We discuss projecting configurations, optimize them and demonstrate their superior performance over DLQR if used alone. Next, we analyze controllable sets and eigenvalues to show advantages and limitations of our proposed controller. Finally, we conclude the paper by discussing potential applications of our proposed architecture.

\section{Discrete DLQR controller} 
\label{sec::discrete}

\begin{figure}[]
        \centering
        \includegraphics[page=1,trim = 0mm 7mm 0mm 0mm, clip, width=0.5\textwidth]{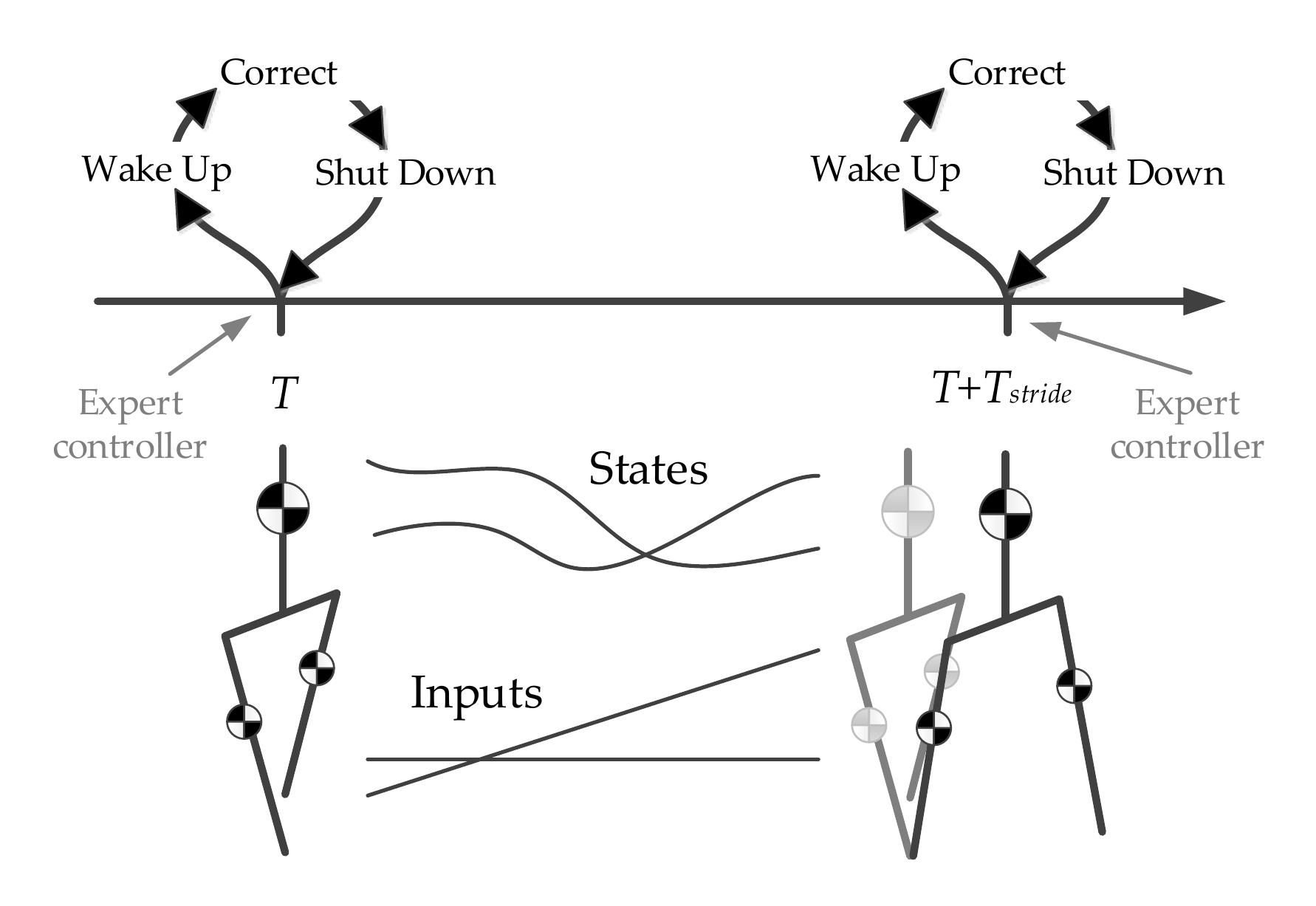}
        \caption{The role of DLQR controller in regulating inputs at the beginning of each phase to stabilize the robot and tracking the nominal periodic solution. After one reaction, it has to wait until the next event to correct for the accumulated error of all disturbances happening in the mid-while.} 
        \label{fig::discrete_controller}
\end{figure}

In 3LP, we consider piecewise linear profiles (constant + time-increasing modes) for each actuator. Linear formulations help us take constants out of the closed form solution and build a linear discrete model. In this model which is described by a transfer matrix $H(T_{stride})$, we can linearly express (in closed form) the state vector after a stride phase (composed of a double support followed by a single support) in terms of the beginning state and inputs. Such formulation allows us to find periodic gaits by imposing symmetry constraints, described in \cite{faraji20163LP}. In this section, we are going to use similar formulations to build a more compact discrete model for an error vector, describing deviations from the nominal periodic solution. In such discrete system which describes error evolution over foot-steps, the error vector is simply calculated in the beginning of each stride phase by some matrix operation. Knowing the error model and effect of inputs, we design a Discrete LQR (DLQR) controller \cite{ogata1995discrete} that finds input modulation to be applied constantly over the whole next stride phase. Such modulation brings the system back to the nominal periodic solution. Figure.\ref{fig::discrete_controller} demonstrates how this controller plays role in our control loop.

\subsection{3LP model}
In this part, we briefly summarize key equations obtained in the first part of this paper about 3LP. Thanks to closed form solutions, one can write:
\begin{eqnarray}
	Q(t) = H(t) Q(0)
	\label{eqn::differentialss}
\end{eqnarray}
Where $H(t) \in \mathbb{R}^{23 \times 23}$ is a combination of individual transfer matrices for single support and double support phases:
\begin{eqnarray}
	\nonumber H(t) = \left\{
	    \begin{array}{ll}
	        H^{ds}(t) & t \le T_{ds} \\
	        H^{ss}(t-T_{ds}) H^{ds}(T_{ds}) & 0 < t-T_{ds} \le T_{ss}
	    \end{array}
	\right.
\end{eqnarray}
Here, $H^{ds}(t)$ and $H^{ss}(t)$ are transfer matrices for double and single support phases with durations $T_{ds}$ and $T_{ss}$ respectively. The vector $Q \in \mathbb{R}^{23}$ represents states and inputs of the model:
\begin{eqnarray}
	\nonumber & Q(t) = \\ &\begin{bmatrix} X(t)^T & \dot{X}(t)^T & P^T & U^T & rU^T & W^T & d^T \end{bmatrix}^T
	\label{eqn::full_vector}
\end{eqnarray}
Where $X(t) \in \mathbb{R}^{4}$ denotes the state of the system (positions of pelvis and swing foot), $P \in \mathbb{R}^{2}$ denotes the position of stance foot, $U \in \mathbb{R}^{4}$ and $rU \in \mathbb{R}^{4}$ denote constant and time-increasing components of input torques respectively (in swing-hip and stance-ankle), $W \in \mathbb{R}^{4}$ denotes external forces and torques on the torso and $d = \pm1$ represents left or right support phases. All these variables have lateral and sagittal components. Other internal and contact forces are depending on these inputs, encoded already in the $H$ matrix. 

\subsection{Zero foot velocity assumption}

Note that the double support phase has a determined timing $T_{ds}$ while the single support phase finishes when both lateral and sagittal components of the foot velocity become zero. Given a fixed timing $T_{ss}$ for single support, this assumption becomes a constraint for controllers to generate proper input torques. We dedicate 2 input channels for satisfaction of zero velocity assumption which yields in the following augmented transfer matrix, valid only for $t=T{stride}$:
\begin{eqnarray}
	\label{eqn::abstraction}
	H' = H - HS_{M_h}^T(S_{\dot{X}_{2}}HS_{M_h}^T)^{-1}S_{\dot{X}_{2}}H
\end{eqnarray}
Where $S_{\dot{X}_{2}}$ selects two rows that describe swing foot velocities $\dot{X}_{2}$. The matrix $S_{M_h}$ also selects two rows associated to constant components of swing-hip torques, dedicated to constraint satisfaction. We use $H'$ hereafter and only consider time-increasing components of swing-hip torques as two controllable channels. Note that the new system is not constrained anymore. 

\subsection{Periodic gaits}

\begin{figure}[]
        \centering
        \includegraphics[trim = 0mm 0mm 0mm 0mm, clip, width=0.25\textwidth]{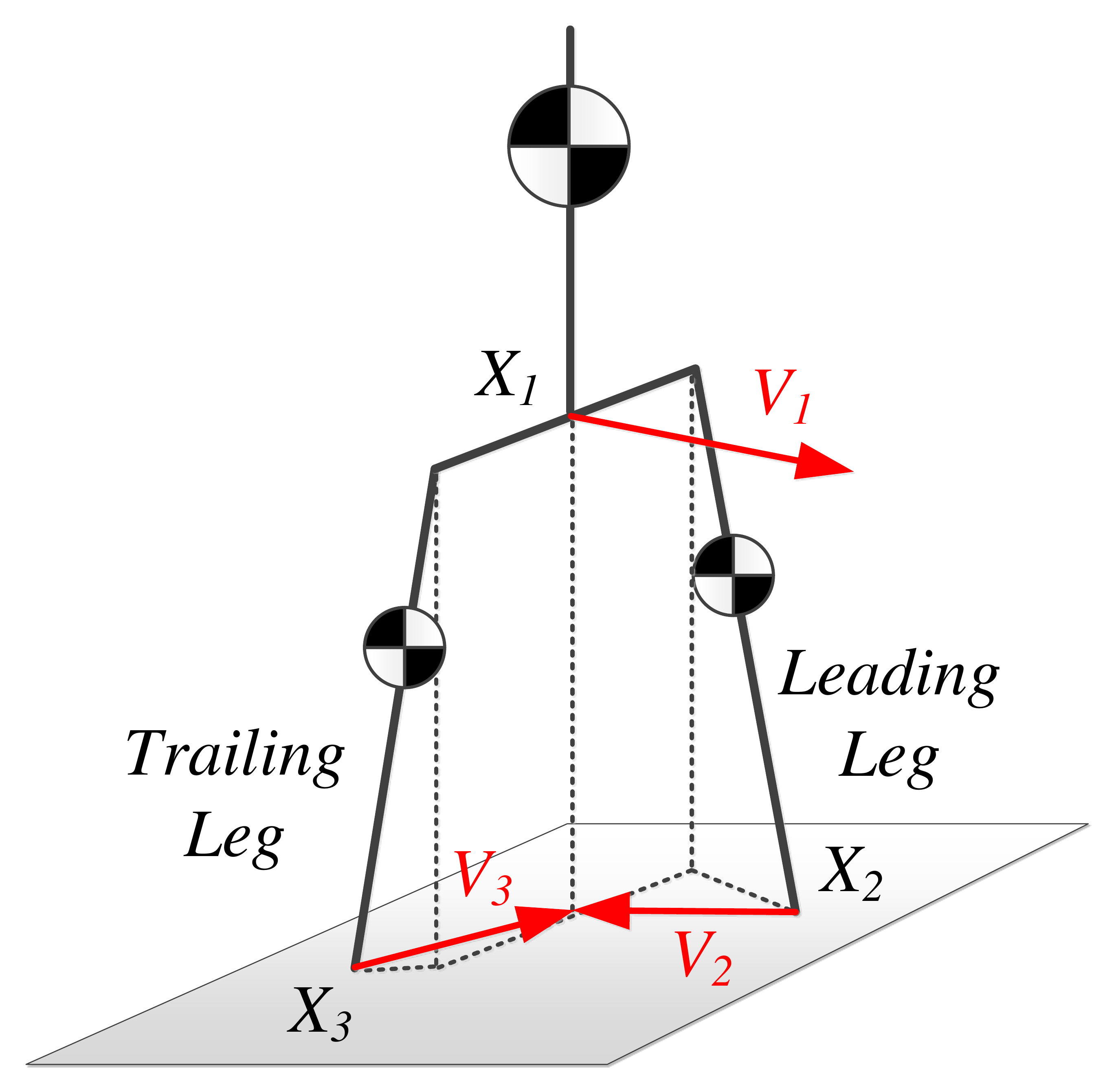}
        \caption{Transforming the global state into local vectors $V_i$ that represent coordinations of the two feet relative to the pelvis (projected on the ground) and the pelvis velocity. All these three vectors have two components in the sagittal and lateral planes.} 
        \label{fig::metric}
\end{figure}

Finding periodic gaits is computationally very easy with 3LP thanks to closed form solutions. We consider three internal vectors $V_i$ representing pelvis velocity and coordinations of the two feet relative to the pelvis. These vectors are demonstrated in Figure.\ref{fig::metric}. For any given state, we can calculate the vectors $V_i$ and compare them to a reference. Such calculation is in fact a local transformation, realized by a simple matrix multiplication. Consider a row selection matrix $S_{XP}$ that selects rows associated to pelvis position $X_1$ and velocity $\dot{X}_1$ and feet positions $X_2$ and $X_3$ (referred to as $P$ in (\ref{eqn::full_vector})). After selecting these 8 rows, the local transformation matrix is:
\begin{eqnarray}
\nonumber M = 
\begin{bmatrix} -1& . & 1 & . & . & . & . & . \\
				. &-1 & . & 1 & . & . & . & . \\
				. & . & 1 & . & . & . &-1 & . \\
				. & . & . & 1 & . & . & . &-1 \\
				. & . & . & . & 1 & . & . & . \\
				. & . & . & . & . & 1 & . & . \\
\end{bmatrix} 
\end{eqnarray}
Note also that after a full stride, one needs the exchange the swing and stance foot locations. This can be simply done by applying the following matrix on the same 8 rows as before:
\begin{eqnarray}
\nonumber T  = 
\begin{bmatrix} . & . & . & . & . & . & 1 & . \\
				. & . & . & . & . & . & . & 1 \\
				. & . & 1 & . & . & . & . & . \\
				. & . & . & 1 & . & . & . & . \\
				. & . & . & . & 1 & . & . & . \\
				. & . & . & . & . & 1 & . & . \\
				1 & . & . & . & . & . & . & . \\
				. & 1 & . & . & . & . & . & . \\
 \end{bmatrix} 
\end{eqnarray}
Considering the fact that after a single stride, we need to revert the lateral quantities, the following matrix $O$ shall be multiplied to the resulting transformed state too:
\begin{eqnarray}
	O = diag([1,-1,1,-1,1,-1])
\end{eqnarray}
Transforming the initial state at the beginning of a stride phase and the one in the end (after $T_{stride}=T{ds}+T{ss}$) provides two vectors of six variables that are equal in a symmetric gait. The transformation operation and imposing symmetry can produce a matrix $R \in \mathbb{R}^{6 \times 23}$ as a function of $H'$:
\begin{eqnarray}
	R = -MS_{XP} + OMTS_{XP}H'
	\label{eqn::periodic}
\end{eqnarray}
All periodic and symmetric gaits with their associated actuation in fact lie in the null-space of $R$, given a specific gait timing. These solutions are initial state vectors represented similar to $Q(t)$ while the $W$ and $P$ components are set to zero. Recall from the first part of this paper that the null-space has multiple dimensions, providing the possibility to find periodic gaits of different actuation inputs. Depending on the choice of $T_{ds}$, there is also a certain single support time where infinite pseudo-passive gaits can be found. These gaits have no hip/ankle actuation and can be simply scaled to modulate the speed. 

\subsection{Discrete error system}

Imagine the nominal periodic solution is a vector $\beta$ with same dimensions as $Q$. The error in the beginning of a stride phase is defined as:
\begin{eqnarray}
	e^- = M S_{XP}(\beta - x^-)
\end{eqnarray}
Where $x^-$ denotes the full state/inputs vector in the beginning of the phase. Since only the states and the $P$ part of $x^-$ are being selected by $S_{XP}$, we can ignore the rest of $x^-$ and break it into individual parts:
\begin{eqnarray}
	e^- =M S_{XP}\beta - M_1X^- -M_2P
	\label{eqn::first_error}
\end{eqnarray}
Where $X^- \in \mathbb{R}^{6}$ denotes the state vector in the beginning of the phase, excluding foot velocity which is zero. $M_1 \in \mathbb{R}^{6 \times 6}$ and $M_2 \in \mathbb{R}^{6 \times 2}$ are matrices coming from the decomposition of $M$. The key formula (\ref{eqn::first_error}) allows to express $X^-$ in terms of $e^-$ as:
\begin{eqnarray}
	X^- = M_1^{-1}(M S_{XP}\beta -e^- -M_2P)
	\label{eqn::first_state}
\end{eqnarray}
We can simply find the state after a full stride phase too. The new error therefore can be written as:
\begin{eqnarray}
	e^+ = M S_{XP}\beta - O M TS_{XP}H' x^-
	\label{eqn::next_error}
\end{eqnarray}
Like before, the quantity $H'x^-$ can be broken into individual matrices:
\begin{eqnarray}
	H'x^- = H' \begin{bmatrix} S_X \\ S_P \\ S_U \\ S_W \\ S_d	\end{bmatrix}^T \begin{bmatrix} X^- \\ P \\ U+U' \\ W \\ d \end{bmatrix}
	\label{eqn::state_evolution}
\end{eqnarray}
Where $U$ is the nominal input torque in the periodic solution and $U'$ is the additional input, produced by the controller. Combining equations (\ref{eqn::next_error}) and (\ref{eqn::state_evolution}), one can write the 6-dimensional error evolution equations as:
\begin{eqnarray}
	\label{eqn::error_full_1}
			 & e^+ = M S_{XP}\beta - O M TS_{XP} H' \times \\
	\nonumber& (S_X^T X^- + S_P^TP + S_U^T (U+U') + S_W^TW + S_d^T d) 
\end{eqnarray}
Replacing $X^-$ from (\ref{eqn::first_state}) in (\ref{eqn::error_full_1}) results in:
\begin{eqnarray}
	\label{eqn::error_full_2}
			  & e^+ = M S_{XP}\beta - O M TS_{XP} H' \times \\
	\nonumber & (S_X^T M_1^{-1}(M S_{XP}\beta -e^- -M_2P) \\
	\nonumber & + S_P^TP + S_U^T (U+U') + S_W^TW + S_d^T d) 
\end{eqnarray}
Now, defining $A=O M TS_{XP} H'$ and $B=S_X^T M_1^{-1}$, one can simplify equations of (\ref{eqn::error_full_2}):
\begin{eqnarray}
	\label{eqn::error_full_3}
			  e^+ &=& M S_{XP}\beta - A \times \\
	\nonumber & &(B(M S_{XP}\beta -e^- -M_2P) \\
	\nonumber & +& S_P^TP + S_U^T (U+U') + S_W^TW + S_d^T d) \\
	\nonumber & =& (AB)e^- + (M S_{XP}-ABM S_{XP}) \beta \\
	\nonumber & +& (ABM_2-AS_P^T)P + (-AS_U^T)(U+U') \\
	\nonumber & +& (-AS_W^T)W  + (-AS_d^T)d 
\end{eqnarray}
Remember from equation (\ref{eqn::periodic}) that the nominal periodic solutions $\beta$ is in the null-space of $R$ matrix. Our error function exactly calculates the mismatch between nominal and current states at the event of touch-down, i.e. beginning of a new stride. Therefore, by removing nominal parts that cancel out, equation (\ref{eqn::error_full_3}) can be reduced to:
\begin{eqnarray}
	e^+ = ABe^- -AS_U^TU' -AS_W^TW
	\label{eqn::error_reduced}
\end{eqnarray}
It should be remembered that the dimension of $U'$ is only 6, losing 2 dimensions due to constraints on the foot velocity. The equation (\ref{eqn::error_reduced}) is useful in the sense that it can predict the effect of disturbances in the system. However, most of the time, disturbances are not known in advance and only happen in a limited time during the stride phase. Therefore, they are not always constant during the whole stride phase, as assumed in the model. We can compensate them in a feed-forward manner if we know them in advance, however in this section we consider no assumption about these disturbances. We only design a controller that observes the state deviation and uses a state feedback to correct it. In the next section however, we deal with intermittent disturbances as well. 

\subsection{DLQR controller}

The design of our controller is in fact straightforward. Although we have 6 actuation dimensions available, we prefer not to modulate the CoP and therefore, only consider time-increasing hip torques which provide 2 actuation dimensions. As motivated in \cite{faraji20163LP}, we leave the CoP modulation authority for low-level controllers to track the abstract motion more precisely. Note that constant hip torques are already dedicated to the foot velocity constraint. Now, an optimal controller could be designed using DLQR routine of MATLAB, which calculates a state feedback matrix $K \in \mathbb{R}^{2 \times 6}$. This controller works only in the event of touch-down. It calculates the error $e^-$, uses the optimal feedback $K$ and produces additional actuation inputs $U'$ which are added to the nominal actuator inputs $U$. The newly adjusted inputs are then constantly applied to the system during the whole following stride phase to stabilize the system. 

We consider three different cost configurations for DLQR design. Keeping the state cost equal to unity matrix, we use input cost matrices of $0.01$ (aggressive), $1$ (normal) and $100$ (light) on the diagonal. Indeed, penalizing inputs by higher cost means less control effort (light controller) and lower cost means fast response (aggressive controller). Throughout the rest of this paper, we always consider these 3 variants to better investigate the behavior of the system. 

\subsection{Forward simulation method}

Referring to the first part of this paper again, thanks to linearity, one can derive $G$ matrices similar to $H$, describing state evolution over time-steps $\Delta t$.
We use $G$ matrix for forward simulation, where in each time-step, the previous state is evolved for $\Delta t$. Although we can directly calculate such evolution using the $H$ matrix and the initial state, we prefer to use $G$ for the sake of visualization and simulation of intermittent disturbances later.

\subsection{Disturbance observer}
Remember that our model by construction, switches to double support when the foot velocity becomes zero. This assumption requires active control of dedicated inputs and previously known terrain profile. Looking at the mismatch of expected and actual state evolution at each time-step, we can estimate the disturbance by pseudo-inversion. Now using $G$ matrices at each time-step, we can calculate the required input for dedicated dimensions which are constant components of hip torques to ensure zero foot velocity in the end. Here the G matrix is of course calculated for the evolution from the current time-step to the end of the phase. Note also that in this work, we assume flat terrain profile and only observe disturbances caused be external forces (and not uneven terrain for example).

\subsection{Numerical results}
To demonstrate the performance of our DLQR controller, we consider three different scenarios of speed tracking, stride-long and intermittent push recoveries. 

\subsubsection{\textbf{Scenario I, Speed modulation:} }

In this scenario, using the simple 1-dimensional null space available for pseudo-passive walking (refer to the first part of this paper), we track different speeds with the same optimal controller. Remember that in case of pseudo-passive walking, an infinite range of speeds could be produced just by scaling the nominal solution in the null-space of $R$ and the motion still remains pseudo-passive. The controller is however general, being able to handle actuated gaits too. The aim of this scenario is to investigate transient conditions.

\begin{figure}[]
        \centering
        \includegraphics[trim = 5mm 8mm 5mm 0mm, clip, width=0.5\textwidth]{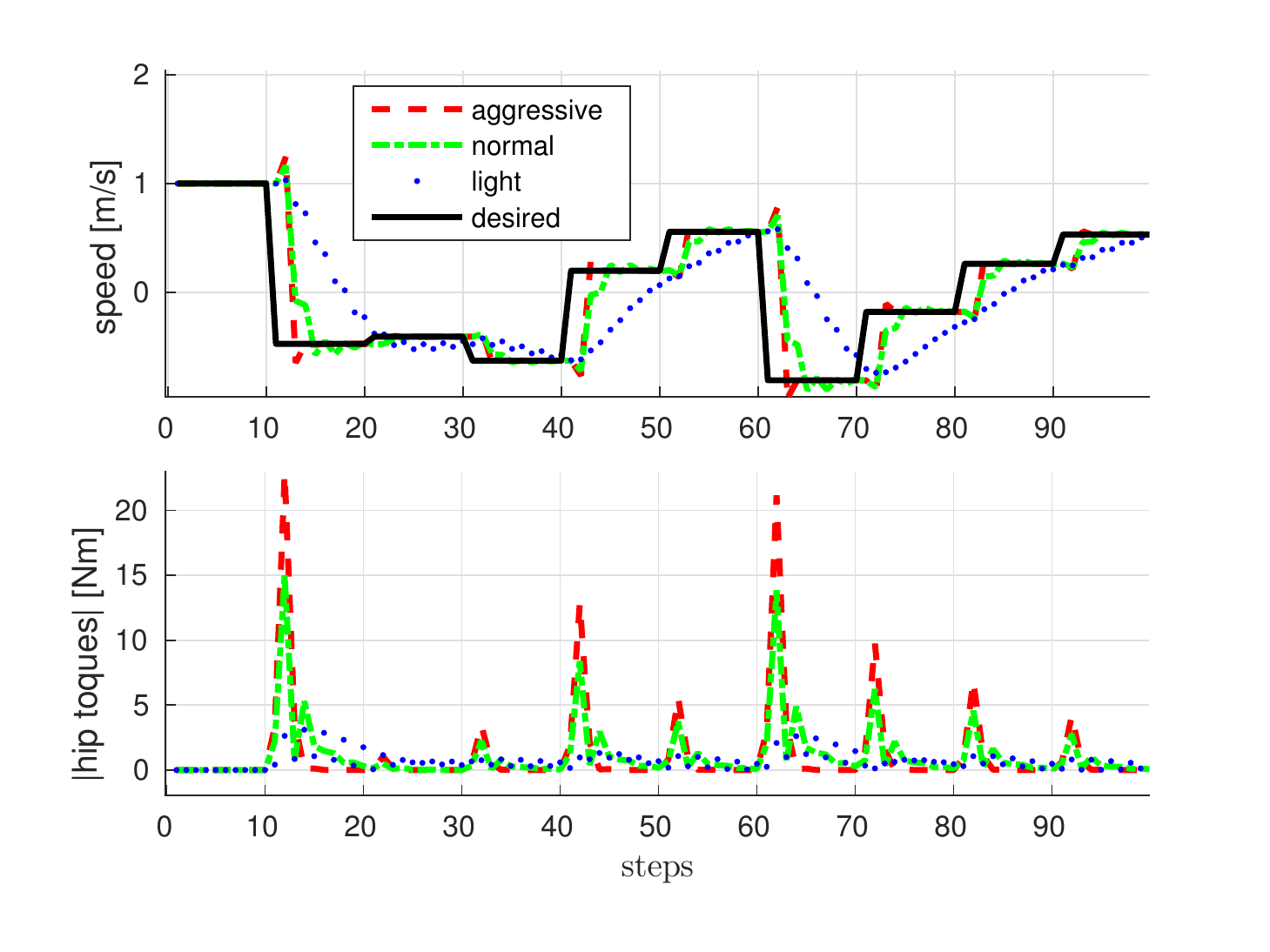}
        \caption{Performance of DLQR controllers in transitioning between different speeds. The original gait is pseudo-passive walking, simply scaled to modulate the speed. Since the velocity of pelvis has continuous-time variations, we measure the actual speed at each discrete event by dividing feet distance with stride time. On the top graph, the desired velocity profile is shown by dashed lines. Other three curves show the performance of aggressive, normal and light controllers. The bottom graph is showing the root-mean-square of all four components of the hip torques. One can clearly see that the aggressive controller performs the transition in almost 2 steps (in accordance with \cite{zaytsev2015two}), but requiring larger inputs. Other two controllers are tracking more smoothly. Movies of this scenario could be found in Multimedia Extension.}  
        \label{fig::tracking}
\end{figure}

In Figure.\ref{fig::tracking}, resulting trajectories of all aggressive, normal and light controllers are demonstrated, given the same desired speed profile. The corresponding movies of this scenario could be found in Multimedia Extension. For any desired speed, one can simply scale a nominal solution (for example for $1m/s$) to calculate a new $\beta$ vector and track it via the DLQR controller. It is obvious from Figure.\ref{fig::tracking} that the aggressive controller is similar to human in requiring almost only two steps to stabilize the walking (\cite{zaytsev2015two}). The normal controller has quite similar performance, but with considerably smaller inputs. The light controller however is very slow in tracking which is not desired. Two important advantages of our method are the simplicity of calculating nominal gaits online and pre-computation of the desired feedback matrix offline. 

\subsubsection{\textbf{Scenario II, Discrete push recovery:} } 

\begin{figure*}[]
        \centering
        \includegraphics[trim = 30mm 5mm 30mm 5mm, clip, width=1\textwidth]{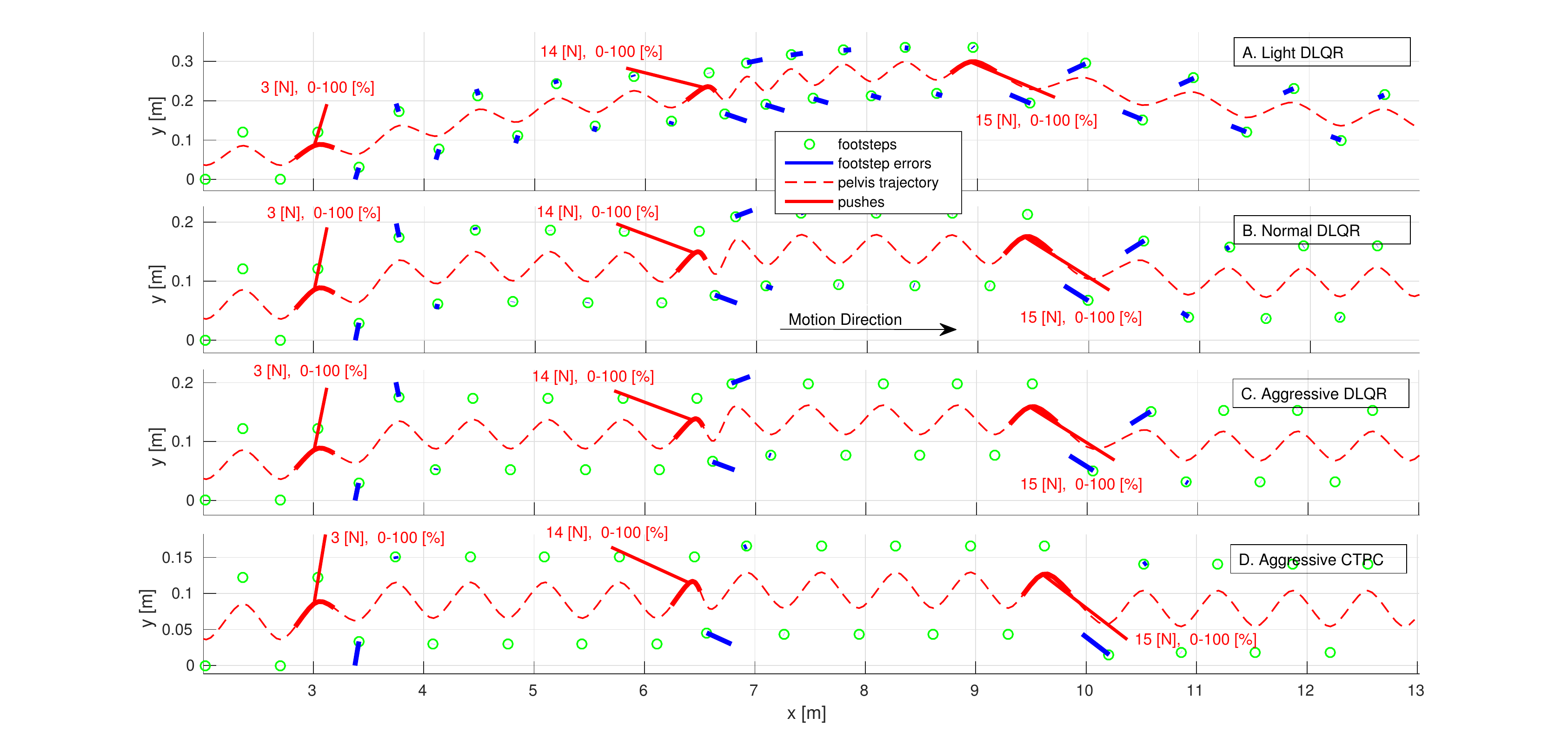}
        \caption{Push recovery of the three DLQR variants (A,B and C) and CPTC (D). Red arrows show external push vectors, red curves show the trajectory of CoM when the push is being applied and blue arrows show relative foot-placement errors. In this figure, the pseudo-passive walking solution is used at $0.5m/s$. All controllers modulate input torques to recover the desired rhythm of motion, but with certain dynamics. \textbf{B,C} The aggressive and normal DLQR controllers take a relatively large corrective step as fast as possible to recover the push. \textbf{A:} The light DLQR controller however has a much longer recovery time. \textbf{D:}  The CTPC controller with C1 architecture takes a large corrective step when the push is being applied and captures most of the error during the same stride phase. In contrast, even the best DLQR controller (aggressive) requires two steps to recover. Movies of this scenario could be found in Multimedia Extension.} 
        \label{fig::full_pushes}
\end{figure*}

In this case, we apply stride-long constant pushes and observe how the model takes corrective steps to stabilize. For simplicity, we push the model only by external forces on the torso and not torques, because they have similar effects and the framework is general enough to consider both at the same time. Since the push is constant over the full stride phase, the closed form solutions and therefore the discrete controller can be used here directly. During the same pseudo-passive walking at $0.5m/s$, we apply different external pushes which are constant in magnitude and direction over the whole stride phase. Resulting trajectories of three controller variants are shown in Figure.\ref{fig::full_pushes}.(A,B and C) while the corresponding movies could be found in Multimedia Extension. All controller variants are able to recover lateral and sagittal pushes with different magnitudes. Similar to the speed tracking scenario, the aggressive controller outperforms the other two controllers, but at the cost of requiring larger inputs. It should be noted that there is no planing or complex controller used to find optimal inputs and the DLQR controller requires minimal calculations online. The disadvantage however is a delayed reaction and the fact that we can not consider step-length and actuation limitations here. 

\subsubsection{\textbf{Scenario III, Intermittent push recovery:} } 

\begin{figure*}[]
        \centering
        \includegraphics[trim = 30mm 0mm 30mm 0mm, clip, width=1\textwidth]{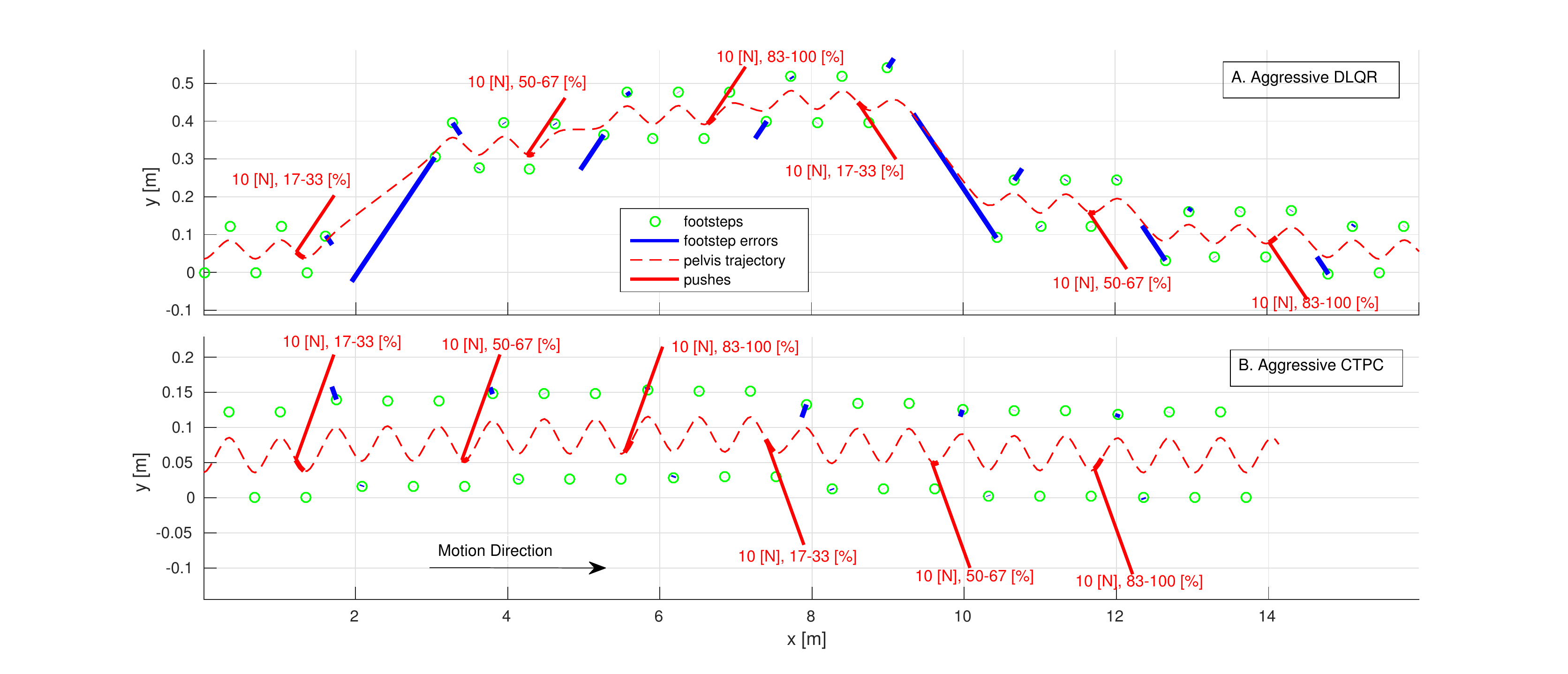}
        \caption{ Performance of the aggressive DLQR (A) and aggressive CTPC (with C1 architecture) (B) controllers in rejecting intermittent external pushes. In this scenario we apply a push with the same magnitude and duration, but at different times during stride phase. We also apply the same pushes in the opposite lateral direction when the other leg is in swing phase and the CoM has a different lateral motion direction. Along with force magnitudes, we show by percentage, the time period in which the push is being applied. \textbf{A:} It is notable that with DLQR controller like Figure.\ref{fig::full_pushes}, the reaction is only taking place after the push and the controller has a delay of course. It is also notable that although pushes have relatively similar magnitudes and much shorter duration compared to Figure.\ref{fig::full_pushes}, they can affect the motion more severely. A push at the end of the phase has relatively small impact. However the same push at the beginning and in the middle of the phase can result in large corrective steps. Also a push towards the swing leg has more severe effect, observed in early-phase pushes more clearly. \textbf{B:} The performance of CTPC is much better, because it can observe the disturbance and react to it in every time-step. Movies of this scenario could be found in Multimedia Extension.} 
        \label{fig::partial_pushes}
\end{figure*}

Although we have a discrete model that can predict the future very fast in terms of computation, it is not always realistic to consider pushes that have synchronous timing with the rhythm of motion. There might be intermittent pushes that shortly act on the robot at any time and disappear. Figure.\ref{fig::partial_pushes}.A demonstrates the resulting trajectory of pseudo-passive walking at $0.5m/s$, subject to intermittent disturbances. Here we consider a push with the same magnitude and duration, but being applied at different times during a stride phase. As expected, late pushes have less impact while early pushes can cause large steps which are not desired. This can be confirmed by the exponential effect of sensory or model errors in forward integration of our equations \cite{bhounsule2015discrete}. With these DLQR controllers, the reaction is taken only in touch-down events which can introduce delay if the push is applied much earlier in the beginning of the phase. Although the model can be stabilized again, large input torques or step lengths are not desired in practice.

\subsection{How to handle intermittent pushes?}

Intermittent disturbances of course break our assumptions in making a discrete model, but they can still be rejected using the available expertise of the DLQR controller. Reaction to intermittent pushes should be done with proper timing however. In the next section, we are going to introduce a computationally simple control law that can stabilize the robot against these realistic pushes as well. Remember that we need a disturbance observer to calculate correct hip torques for the foot velocity constraint, as explained before.

\section{Continuous-Time Projecting Controller (CTPC)}

Although we found a closed form solution to the state evolution equations, we still need to have an active controller at each time-step during the stride phase. The goal is to react with minimal delay against intermittent disturbances which are estimated by our disturbance observer continuously. With such information, we can use the expertise of DLQR controller to find optimal control inputs for the current time-step that stabilize the robot and of course satisfy the foot velocity constraint. 

\subsection{The core idea}

\begin{figure}[]
        \centering
        \includegraphics[page=2,trim = 3mm 10mm 3mm 0mm, clip, width=0.5\textwidth]{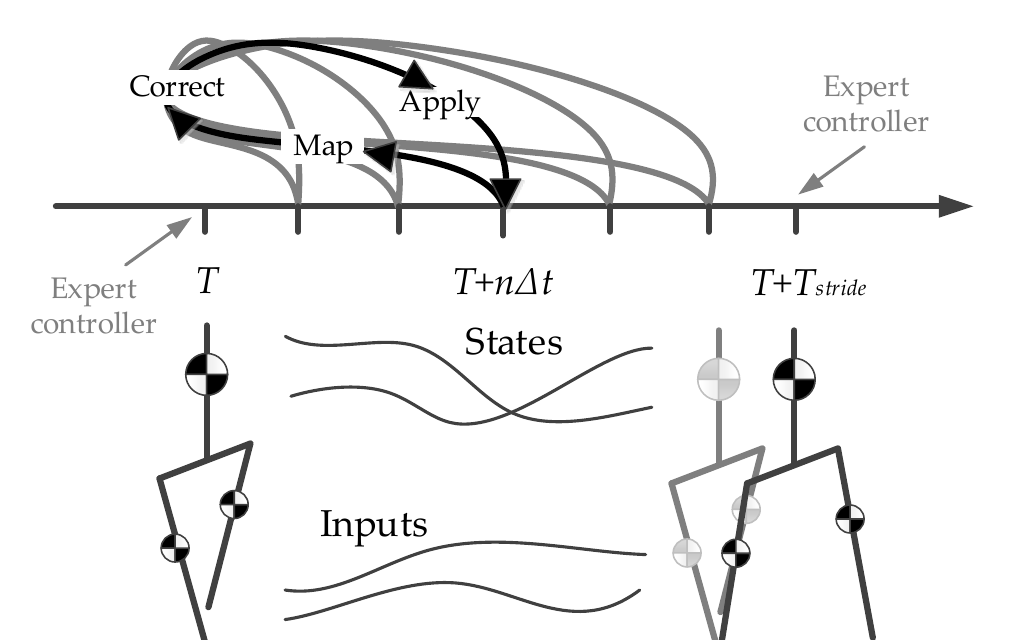}
        \caption{A demonstrative schematic of the idea behind time-projection. This controller in fact relies on the expertise of DLQR controller designed earlier. At each time-step $T+n\Delta t$, the observed state is mapped to the beginning of the phase, optimal inputs are calculated using DLQR controller and then, these inputs are used at the current time step. One can of course expect that depending on the controller and the nature of disturbances, input profiles might not be constant or linearly increasing with time like Figure.\ref{fig::discrete_controller}. } 
        \label{fig::cont_controller}
\end{figure}

At each time step, after disturbance estimation, we project the error back in time to the beginning of the phase. The projected error can now be corrected by the expert DLQR controller designed before. The resulting optimal inputs are then used at the current time step to stabilize the system. Figure.\ref{fig::cont_controller} demonstrates how we setup our CTPC controller. This new method is simple to execute and it does not need redesign of controller, offline optimizations or advanced MPC frameworks. Similar back projection idea is also used in \cite{byl2008approximate} where a post-collision state should be rewound to a pre-collision state where an additional impulse should be added. Then the system is forward simulated again to find the contribution of the additional impulse. Here we use projection at any time however, conceptually shown in Figure.\ref{fig::cont_controller}. 

\subsection{How to realize the time-projection}

The idea of projection is to find evolution of the currently observed state $X_t$ (we call it current system hereafter) and some unknown initial states $X_i$ (which are called alternative systems) until the end of stride phase by $G$ and $H$ matrices respectively and force them to be equal. Effectively, this is a projection back in time to find the cause (inputs and initial states) of the error observed in the current state. However, this projection is done through looking into future (the end of stride phase), to ensure satisfying the zero foot velocity constraint for the current and alternative systems. Such constraint indeed requires prior knowledge of the disturbance profile, because the controller needs to modulate actuation torques properly to ensure zero foot velocity at the end of stride. In this work for simplicity, we assume that the currently observed disturbance will remain constant until the end of the stride, which let us use the transition matrices calculated in \cite{faraji20163LP} easily. In case a prior knowledge of the disturbance profiles is available, one can flexibly change the controller formulations, though the whole generic design process remains the same.

Given the whole idea, how should we determine the inputs of alternative systems and whether disturbances influence them or not? We are only looking for an optimal policy to be applied at the current time-step. However the alternative systems are in fact working in closed-loop as well, meaning that they produce control policies $U_i$ according to DLQR law. The optimal policies $U_i$ are linear functions of their unknown initial states $X_i$. Together with state evolution laws, these equations form a linear system of $Ax=B$ that can be solved very fast at each time step. Here, unknowns are initial states and of course their associated control policies.

\subsection{Important features}

\begin{figure}[]
        \centering
        \includegraphics[trim = 10mm 5mm 10mm 5mm, clip, width=0.45\textwidth]{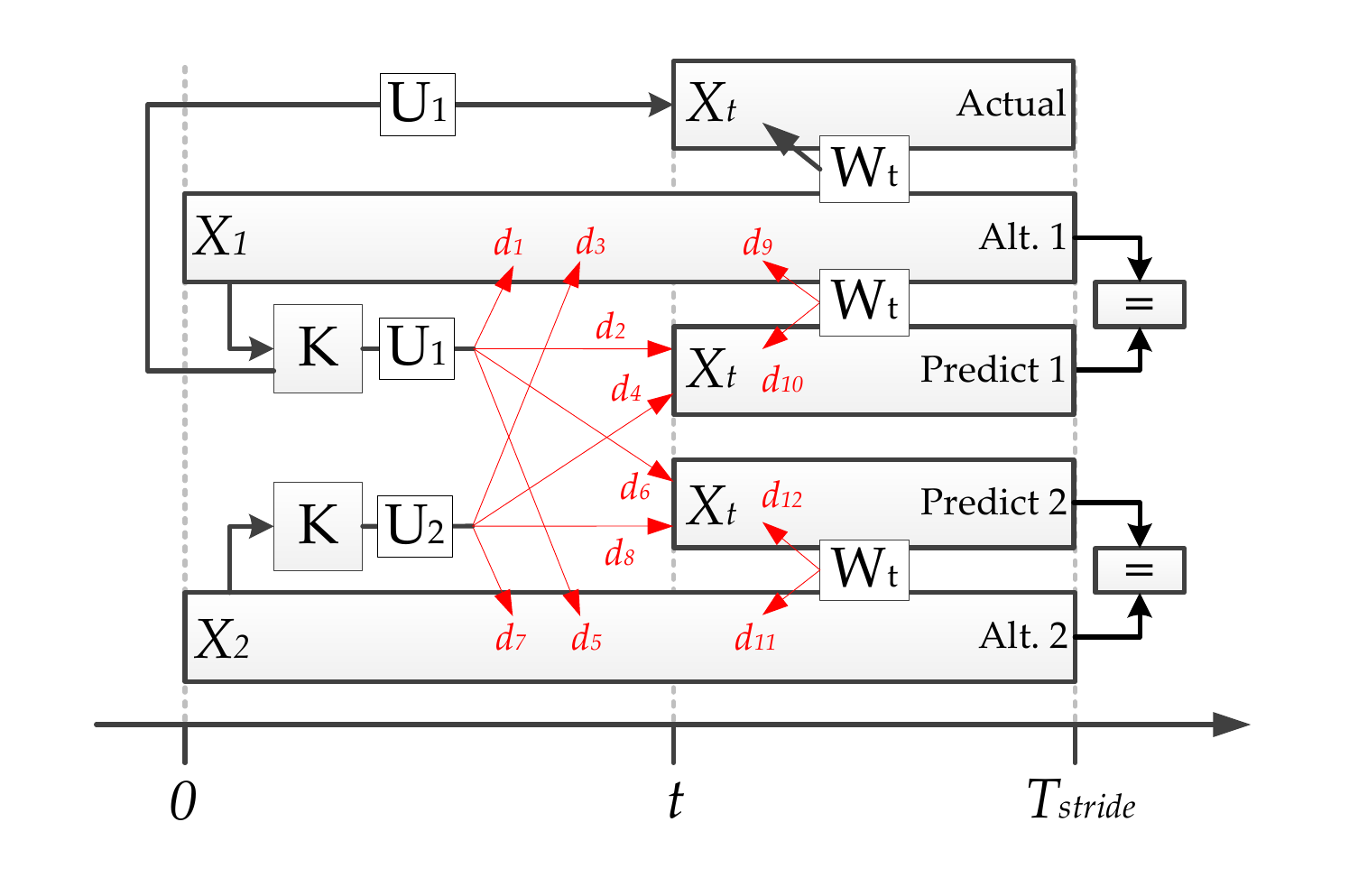}
        \caption{The proposed continuous-time control architecture with unknown interconnections shown in red. This diagram shows how the current state $X_t$ is projected back in time to produce two unknown initial states, each generating a feedback according to the DLQR matrix. These initial states are evolved until the end of the same phase, when they produce the same state as the evolution of $X_t$ does. Feedback inputs as well as disturbances at current time $W_t$ could influence any of alternative or prediction systems. Then, the feedback $U_1$ produced by the first alternative system is given to the current system which is of course subject to disturbance. Note that variables to be known when using this architecture are $X_t$ and $W_t$ at each instance of time and the resulting output is $U_1$. } 
        \label{fig::cont_arc}
\end{figure}

There are two features very important to investigate:
\begin{itemize}
	\item The first feature is indeed the number of alternative systems and the interconnections between them. Each alternative system produces an optimal policy which can be used in itself or other systems. More alternative systems can therefore help to decouple dynamics of different kind, for example system dynamics from disturbance dynamics. We limit this study to considering one or two alternative systems only.
	\item The second feature is whether the controller produces constant inputs or not. In other words, with stride-long constant disturbances, could the same control policy be produced only once in the beginning of the phase and repeated afterwards? This is basically equivalent to another discrete controller, superior to DLQR, because it can handle intermittent pushes and the problem of timing too. In terms of computation also, this controller is more efficient as it becomes active only when a new disturbance is observed.
\end{itemize}
These two features produce four different categories of architectures, C1 to C4 (refer to Table.\ref{table::controllers}). Besides, the control policies produced by each alternative system can be given to others as well. Due to arbitrary coupling of alternative systems therefore, we setup a general set of equations for the case of two alternative systems and later specify conditions which lead to the desired features mentioned. A more demonstrative schematic of projecting architecture is shown in Figure.\ref{fig::cont_arc}. 

\subsection{Mathematical formulation}
Because of zero foot velocity assumption, again we use $H'$ and $G'$ matrices to make sure we always satisfy this constraint for the current and alternative systems. The matrix $G'$ is derived from $G$ similar to (\ref{eqn::abstraction}). Like previous section, we select time-increasing hip torque components as active control input and do not modulate the CoP. 

As stated before, initial states and inputs of the alternative systems are unknown, though stance foot location and $d$ variables are known. Additionally, we are going to modulate control inputs, hence nominal gait actuations are still being considered. By decomposing $H'$ into different blocks, we can write initial state evolution as:
\begin{eqnarray}
	X_i(T_{stride}) = H_1X_i(0)+H_2U_i+H_3W+H_4Z
	\label{eqn::cont1}
\end{eqnarray}
Where the variable $X_i\in \mathbb{R}^6$ denotes the initial state except foot velocity which is zero, $U_i\in \mathbb{R}^{2}$ denotes the time-increasing hip torque component, $W\in \mathbb{R}^{4}$ denotes external disturbing force (coming from state estimator) and $Z\in \mathbb{R}^{11}$ contains all other constant terms like stance foot position $P$, nominal hip/ankle torques and the variable $d$. Matrices $H_i$ are coming from the 6 equations describing evolution of the state variables except foot velocity. Similarly, the same decomposition could be derived for the current state $X_t$ using $G'$, describing state evolution from the current time to the end of the phase:
\begin{eqnarray}
	X_t(T_{stride}) = G_1X_t+G_2U_t+G_3W+G_4Z
	\label{eqn::cont2}
\end{eqnarray}
Where $X_t\in \mathbb{R}_8$ denotes the current full state vector, including foot velocity. Remember that foot velocity is not necessarily zero in the middle of the phase. 

Apart from 6 state equations, there are other 2 equations for each alternative system that relate the initial state to the control input using DLQR law:
\begin{eqnarray}
	U_i = -K e_i = -K (M S_{XP}\beta - M_1X_i -M_2P)
\end{eqnarray}
Where $e_i\in \mathbb{R}_{6}$ represents the error of initial state $X_i$ with respect to the nominal solution $\beta$. This equation can be shown in abstract form by defining auxiliary $D$ matrices:
\begin{eqnarray}
	U_i + D_1X_i = D_2 + D_3P
	\label{eqn::cont3}
\end{eqnarray}
Using equations (\ref{eqn::cont1},\ref{eqn::cont2},\ref{eqn::cont3}) and the fact that $X_i(T_{stride})=X_t(T_{stride})$, we can setup our linear system of equations, representing 2 alternative systems and the current system:
\begin{eqnarray}
	\nonumber & \begin{bmatrix}
		H_1 & d_1H_2-d_2G_2 & 0 & d_3H_2-d_4G_2 \\
		D_1 & I & 0 & 0 \\
		0 & d_5H_2-d_6G_2 & H_1 & d_7H_2-d_8G_2 \\
		0 & 0 & D_1 & I
	\end{bmatrix} \begin{bmatrix}
	X_1 \\ U_1 \\ X_2 \\ U_2
	\end{bmatrix} = \\ & \begin{bmatrix}
	G_1X_t+(d_{10}G_3-d_{9}H_3)W+(G_4-H_4)Z \\
	D_2 + D_3P \\
	G_1X_t+(d_{12}G_3-d_{11}H_3)W+(G_4-H_4)Z \\
	D_2 + D_3P \\
	\end{bmatrix}
\end{eqnarray} 
In this set of equations, initial states $X_1$ and $X_2$ of the two alternative systems produce $U_1$ and $U_2$ respectively where $U_1$ is given to the current system. Binary variables $d_i$ where $1 \le i \le 12$ indicate whether inputs or disturbances are used in alternative and current systems or not. At each instance of time, the variables $X_t$, $W$ and $Z$ are known. However initial states and their associated feedback inputs are unknowns to be found. Figure.\ref{fig::cont_arc} shows a graphical demonstration of the proposed continuous-time control architecture with unknown interconnections. The most complicated version with both alternative systems results in a $16 \times 16$ set of equations $Ax=B$ which can be solved in few micro-seconds using modern linear Algebra libraries. Also, the DLQR matrix can be calculated off-line as well as all model matrices like $H$ and $G$.

\subsection{Search for the best configuration}

In this section, we are going to search over $2^{12}=4096$ different configurations of the binary variables $d_i$ for the best performance. In case of having only a single alternative system, it might be easy to qualitatively argue about the limited number of configurations. However for two alternative systems, it becomes complicated to propose configurations based on control insights. The complex architecture of CTPC can embed optimum stabilizing policies that cannot be determined with intuition easily. Note that the DLQR controller is already optimal and stable, but we are going to use it as a seed in the continuous-time architecture. Therefore, it is not easy to determine qualitatively if a candidate configuration is stable or optimal anymore. 

\begin{figure}[]
        \centering
        \includegraphics[trim = 10mm 0mm 10mm 0mm, clip, width=0.5\textwidth]{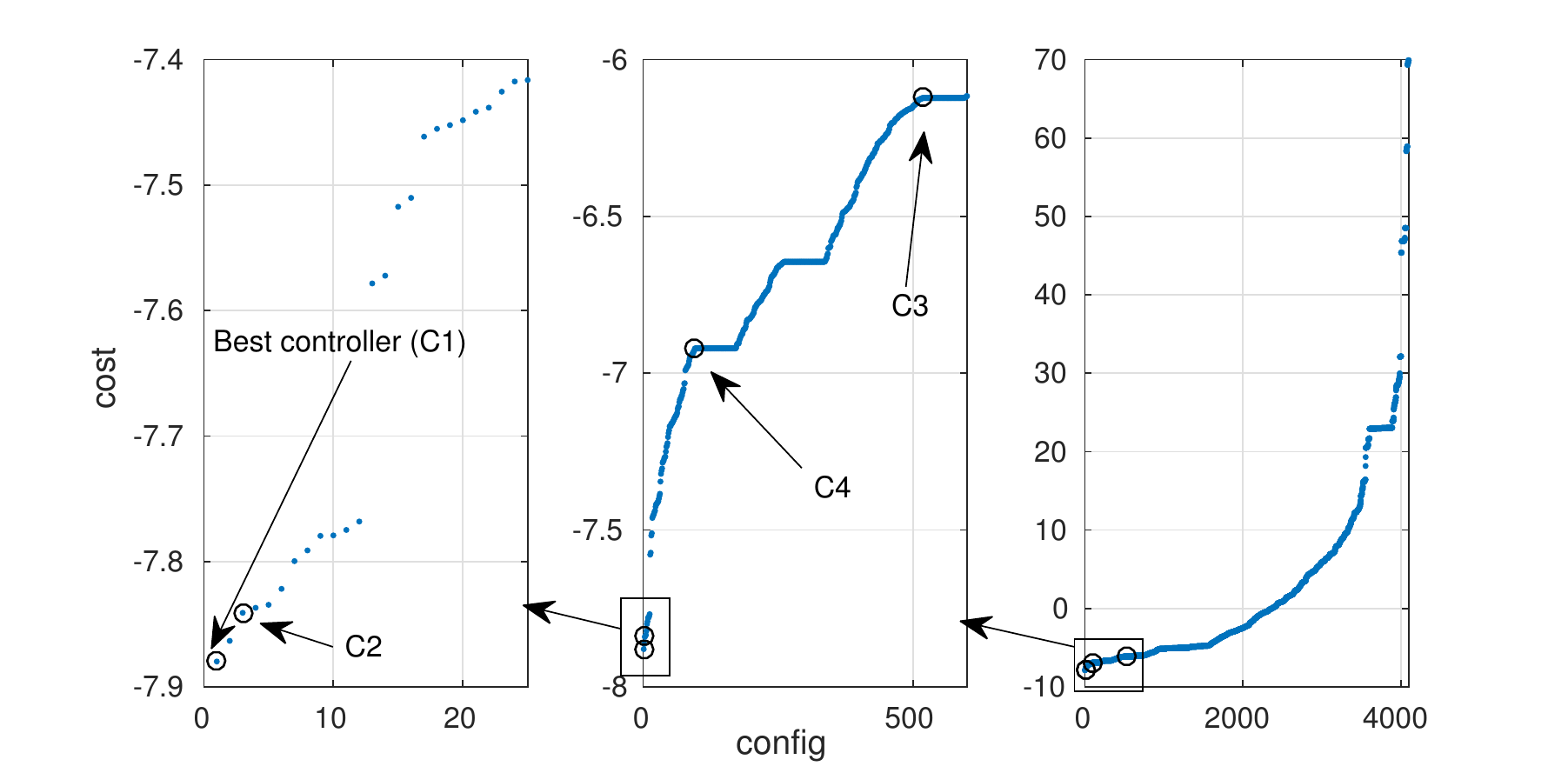}
        \caption{The sorted cost of all configurations, shown in logarithmic scale. C1 and C2 are very similar and provide superior performance compared to C3 and C4, because of exploiting two alternative systems. C3 and C4 however have much higher costs, because of single-alternative structure.} 
        \label{fig::costs}
\end{figure}

In this paper, we propose a systematic search process based on intrinsic properties of the projecting architecture. We basically consider a finite-horizon quadratic cost similar to DLQR to quantify the cost of rejecting disturbances with different timings. The search process over $1 \le c \le 2^{12}$ configurations has the following steps:
\begin{itemize}
	\item \textbf{Model Geometry:} The procedure is done for two adult-size and kid-size models $m \in \{A,K\}$. This will make the controller more robust against model properties.
	\item \textbf{DLQR selection:} The procedure is also done for all three variants of DQLR matrix with $Q=1$ and $R_1=10^2$ (light), $R_2=1$ (normal) and $R_3=10^{-2}$ (aggressive) matrices. $Q$ and $R_k$ here represent state and inputs cost matrices used in DLQR design.
	\item \textbf{Self-stability:} We consider 6 initial state vectors, each perturbed in one dimension. Each vector is simulated over $N=10$ steps and the sum of state/input costs are calculated for $1\le i\le6$:
	\begin{eqnarray}
		^m_kS_c^i = \begin{bmatrix} \sum_{1}^{N} e^Te & \sum_{1}^{N} U'^TR_kU' \end{bmatrix}
	\end{eqnarray}
	\item \textbf{Perturbations:} We also split the whole $T_{stride}$ into $n=7$ sub-periods and consider ${n \choose 2}$ pushes of same magnitude, but different timing. Each push starts from $0 \le j_1 \le n-1$ sub-period and finishes in $j_1+1 \le j_2 \le n$ sub-period. Similar state/input costs are calculated then for each $1\le j\le{n \choose 2}$ cases with weighting coefficients proportional to the timing:
	\begin{eqnarray}
		^m_kW_c^j = \begin{bmatrix} \sum_{1}^{N} e^Te & \mu (j_2-j_1)^2 \sum_{1}^{N} U'^TR_kU' \end{bmatrix}
	\end{eqnarray}
	Here $\mu=10^{-2}$ makes the architecture more aggressive in rejecting pushes while self-stability is kept untouched.
	\item \textbf{Cost vector:} Now for each configuration $c$ and DLQR variant $k$, we concatenate all the cost vectors calculated before:
	\begin{eqnarray}
		_kV_c = \Omega_{m=\{A,K\}} \begin{bmatrix} \Omega_{i=1}^6 [^m_kS_c^i] & \Omega_{j=1}^{n \choose 2} [^m_kW_c^j] \end{bmatrix}
	\end{eqnarray}
	Where the operator $\Omega$ represents row-wise concatenation of row-vectors.
	\item \textbf{Normalization:} Now for each dimension in the cost vector, the minimum $_kU$ over all configurations is found. Then, all cost vectors $_kV_c$ are normalized by the inverse of these minimum values to get $_k\hat{V}_c$.
	\begin{eqnarray}
		_kU = \underset{c}{\text{min}} ({}_kV_c) \rightarrow {}_k\hat{V}_c = log_{10}\frac{_kV_c}{{}_kU}
	\end{eqnarray}
	Where the division and $min$ operators are calculated element wise.
	\item \textbf{Final cost:} The final cost of each configuration $c$ is now calculated as follows:
	\begin{eqnarray}
		V_c = \sum_{k=1}^3 ||{}_k\hat{V}_c||_1
	\end{eqnarray}
	\item \textbf{Selection:} In each of four categories C1 to C4, the configuration with minimum cost is selected.
\end{itemize}
The proposed search process considers model variations, DLQR design, intermittent push timing and self-stability criteria all together to find the optimal yet general projecting architecture. After the search process, we have a sorted list of costs shown in Figure.\ref{fig::costs}. As previously mentioned, we are specially interested to know how advantageous it is to use two alternative systems instead of one and also if control inputs are constant or not. For the first feature, we can simply find solutions with  $d_3=d_4=0$ (Figure.\ref{fig::cont_arc}). For the second feature, we numerically simulate a single stride with constant initial and intermittent disturbance to see if inputs stay constant during the whole stride or not.

\begin{figure}[]
        \centering
        \includegraphics[trim = 5mm 5mm 5mm 0mm, clip, width=0.5\textwidth]{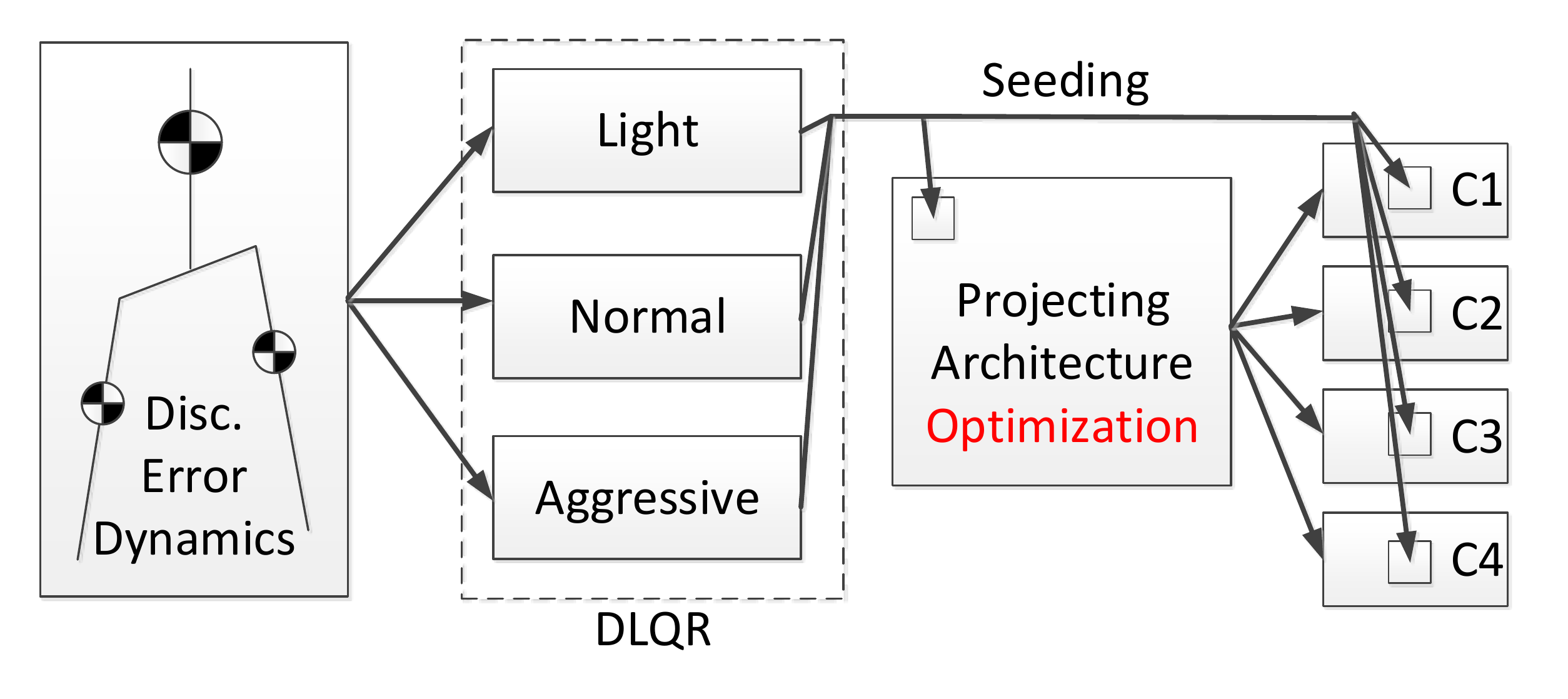}
        \caption{An overview of the search process for finding the best projecting configuration. All three variants of DLQR controllers serve as seeding matrices in CTPC.} 
        \label{fig::optim_proc}
\end{figure}

An overview of the search process is demonstrated in Figure.\ref{fig::optim_proc}. Based on 3LP discrete error dynamics, 3 variants of DLQR controllers are calculated. These variants are then used as a seeding expert in the optimization of the projecting configuration. For each of four categories regarding the two important features (number of alternative systems and constant inputs), the best configuration is selected (C1 to C4). These configurations as well as the DLQR of previous section are listed in Table.\ref{table::controllers}. Note that all three variants of aggressive, normal and light matrices can be used in C1 to C4 architectures, interchangeably.

\begin{table}[]
  \centering
  \begin{tabular}{ccccc}
  \hline
	Cont. & type & alt. & const. inputs & $d_1 ... d_{12}$ \\
	\hline
	C1 & cont. & 2 & no & 1110 0110 1001 \\
	C2 & cont. & 2 & yes & 1110 0110 1101 \\
	C3 & cont. & 1 & no & 1100 1010 1001 \\
	C4 & cont. & 1 & yes & 1100 1011 1101 \\
	DLQR & disc. & 0 & yes & - \\
  \hline \\
  \end{tabular}
  \caption{Features of the DLQR controller and four best CTPC controllers of different categories (C1-C4) found in this section. }
  \label{table::controllers}
\end{table}

It should be noted that the search process could be alternatively done by simulation of different random pushes. However our method is deterministic and more general. Including the kid-size model which has a different natural frequency due to the smaller size and all DLQR variants help to obtain more robust configurations. We have also numerically confirmed that the cost values are independent of walking speed, for all control architectures. This is due to the fact that error dynamic equations are independent of nominal solutions.  

\subsection{Performance comparison}

\begin{figure}[]
        \centering
        \includegraphics[trim = 12mm 10mm 5mm 5mm, clip, width=0.5\textwidth]{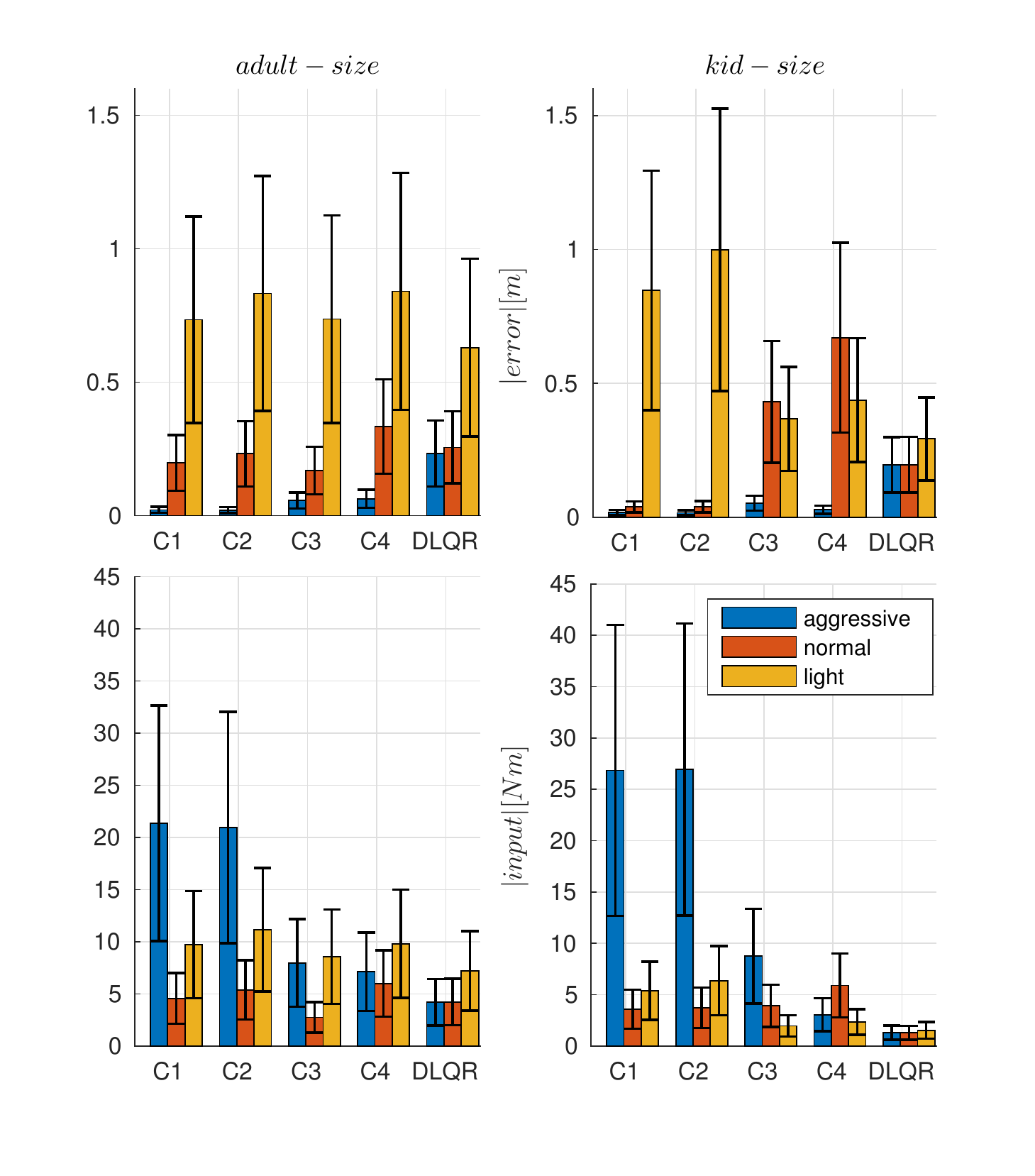}
        \caption{Performance of all 5 controllers (C1 to C4 and DLQR) during a forward simulation of 1000 steps with 100 random external pushes of different magnitude ($\pm 20 N$) and timing (start time and duration). In this plot we show the average magnitude of state errors (\ref{eqn::error_reduced}) and hip-torques for the two different-sized 3LP models.  } 
        \label{fig::peformance}
\end{figure}

Although C1 to C4 are already sorted, we would like to show their performance in a long scenario of walking with random intermittent pushes. Here we quantify the performance by average (per step) magnitude of state errors $e$ and additional hip-torques $U'$ (\ref{eqn::error_reduced}). The resulting statistics are shown in Figure.\ref{fig::peformance} where all architectures are tested with the three matrix variants over adult-size and kid-size models. Remember that DLQR is only acting in the beginning of the phase, while other CTPC controllers update the inputs at any time-step.

\subsubsection{\textbf{The role of seeding matrices:}}

Controllers seeded with aggressive matrix are always performing better, though requiring larger inputs. It is remarkable that all four categories of CTPC outperform DLQR, when seeded with aggressive matrix. The magnitude of inputs required by aggressive matrices are large because of fast reaction. Light matrices also produce large input magnitudes, because the push recovery takes multiple steps. Normal matrices however provide a better compromise, despite producing larger errors compared to aggressive matrices. 

\subsubsection{\textbf{Exploring features:}}

Apart from the seeding DLQR matrix, we are also interested to compare architectures of different features and for this purpose, we focus on the normal DLQR matrix. As expected, using two alternative systems (C1 and C2) is better than one (C3 and C4), verified on both models with different sizes. Controllers with variable inputs (C1 and C3) are also outperforming those with constant input (C2 and C4). However in case of kid-size model, C4 with only one alternative system has quite similar performance with C1 which is the best. Despite being model-dependent, it is surprising that a simpler controller with less computations can give similar performance. Using CTPC, specially the time-varying two-alternative architecture (C1), results in much better performance compared to DLQR. This can be explained by the fact that C1 reacts to disturbances more quickly. 

\subsubsection{\textbf{Footstep plans:}}

To visualize the performance of CTPC, we have demonstrated the previous scenario of Figure.\ref{fig::partial_pushes}.A in Figure.\ref{fig::partial_pushes}.B using CTPC (C1 architecture with aggressive matrix). Although same disturbance pattern is applied to the system, just by finding proper hip-torques and therefore footstep adjustments, CTPC is able to recover with minimal error. Movies of this scenario could be found in Multimedia Extension.

It is also important to compare the recovery of pushes with long duration like Figure.\ref{fig::full_pushes}.(A to C). Using CTPC, we have simulated similar pushes and plotted the curve in Figure.\ref{fig::full_pushes}.D. Again, CTPC outperforms DLQR thanks to the online reaction. We can also see that CTPC recovers the push in the same stride while DLQR shows a delay of at least one step. This indicates that CTPC can solve the timing problem very well. In the next section, we are going to explore other properties of the proposed projecting architecture. We compare it to the DLQR controller in terms of stability and controllable regions. We would also like to predict how much better a more complicated controller like MPC can do, if inequality constraints on actuator limits or footstep locations are considered.

\section{Further analysis of the projecting architecture}

Having optimized projecting architectures, we analyze and compare controllable regions and eigenvalues of the closed-loop systems. In this work, we do not setup MPC controllers. However through the analysis of controllable regions, we will argue on the effectiveness of other controllers including MPC, compared to our proposed architecture. Throughout this section, we take the C1 architecture with aggressive seeding matrix. 

\subsection{Closed-loop eigenvalues}

This analysis quantifies the self-stabilization property of our proposed controller. As demonstrated previously in Figure.\ref{fig::full_pushes} and Figure.\ref{fig::partial_pushes}, CTPC outperforms DLQR in recovering both stride-long and intermittent pushes. In terms of self-stabilization however, CTPC might have different properties. As mentioned before, the analysis of error dynamics and all the eigenvalues calculated in this part are independent of the nominal gait. In other words, they are independent of the type of gait (in terms of actuation) and the speed. This keeps the analysis universal thanks to linearity of 3LP. However due to fixed timing, one needs to consider the effect of stride duration in analyzing stability. In the first part of this paper, we already mentioned how well 3LP can predict the natural relation of velocity and frequency in humans. In this work however we setup all controllers based on fixed timing to keep everything linear. In future works, we will setup nonlinear controllers to adjust the timing as well as considering inequality constraints.

\begin{figure}[]
        \centering
        \includegraphics[trim = 0mm 5mm 0mm 0mm, clip, width=0.5\textwidth]{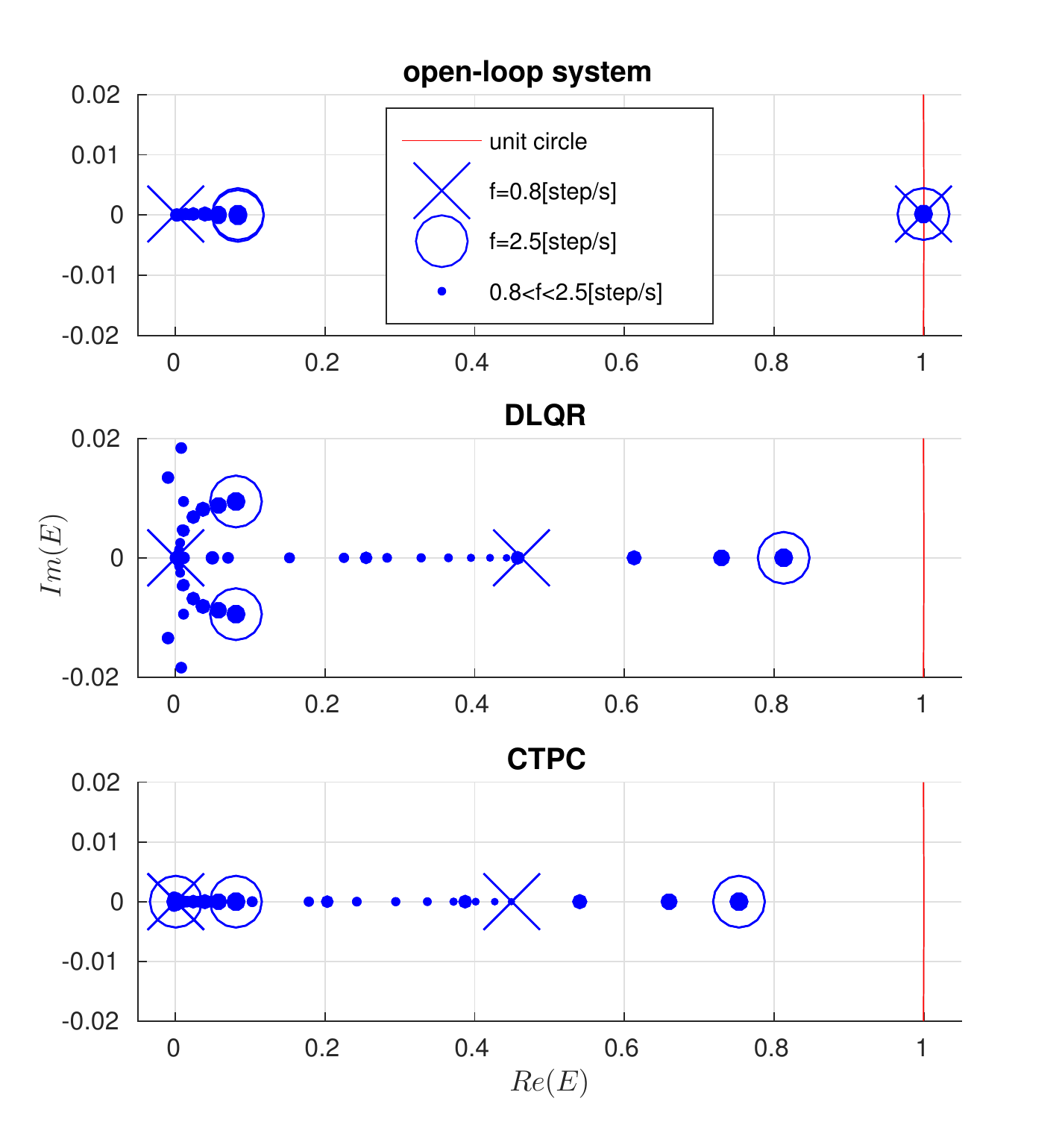}
        \caption{Eigenvalues of the 6-dimensional closed-loop system, calculated over 2 steps. Due to the similarity of lateral and sagittal dynamic equations, eigenvalues appear as 3 duplicates. This figure only shows 4 eigenvalues for the open-loop system, however the other two eigenvalues are large but real, ranging from $12.1$ to $1172$ for walking frequencies of $2.5$ to $0.8 step/s$ respectively. Such range of frequencies is inspired by adult-size human walking data \cite{bertram2005constrained}. In this figure, larger circles show faster frequencies while the slowest frequency is shown by a cross sign.} 
        \label{fig::eigen}
\end{figure}

In this part, we calculate the eigenvalues of the open-loop and closed-loop systems for CTPC and simple DLQR controllers. We fix the double support time to $20\%$ of the whole stride phase and calculate eigenvalues for different walking frequencies shown on Figure.\ref{fig::eigen}. The fixed timing assumption indeed decouples lateral and sagittal dynamics. Due to similarity, lateral and sagittal eigenvalues are the same and one effectively observes 3 duplicates instead of 6 for the full system. In open-loop case, there are 2 eigenvalues always on the unit circle, 2 very large but real (not shown in Figure.\ref{fig::eigen}) and 2 close to zero. Both controllers are stable however. Depending on design cost matrices, the DLQR controller introduces complex eigenvalues while CTPC always shows real eigenvalues, seeded with the same DLQR matrix. CTPC is slightly stronger in self-stabilization based on smaller magnitudes of eigenvalues and the fact that it always has 2 eigenvalues very close to zero. 

Another outcome of eigenvalue plots in Figure.\ref{fig::eigen} is the fact that walking with faster frequencies can be stabilized in fewer steps, shown by smaller eigenvalues. The trade-off is in larger control inputs however required to perform the swing motion, which is more difficult for the real hardware to track. One possible choice is to adjust the frequency according to the desired speed, similar to humans for example \cite{bertram2005constrained}. In terms of control, changing timing results in nonlinear formulations which remains out of the scope of this work.

\subsection{Push recovery strength}

\begin{figure*}[]
        \centering
        \includegraphics[trim = 30mm 5mm 20mm 0mm, clip, width=1\textwidth]{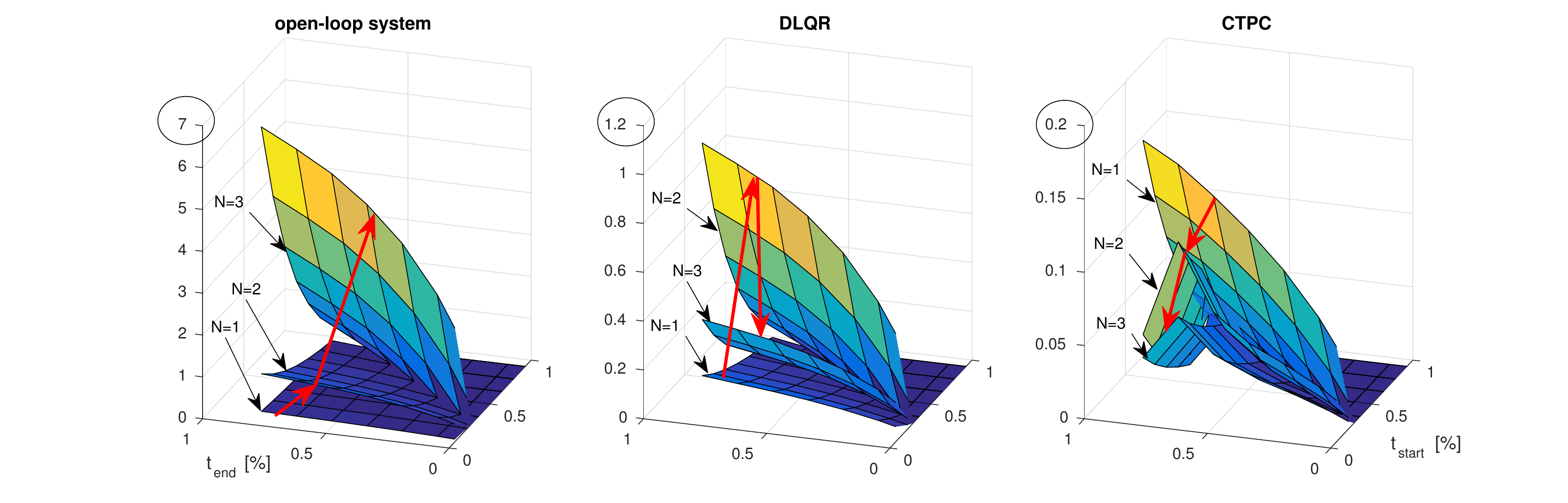}
        \caption{Demonstration of intermittent push recovery performance for the open-loop system, DLQR feedback and CTPC (C1 with aggressive matrix). In this plot, we inspect the effect of start and finish times as percentage of the stride phase (refer to Figure.\ref{fig::partial_pushes}). Surfaces show the norm of 6-dimensional error, calculated at touch down events over three consecutive. The open-loop system is unstable, DLQR controller overshoots and CTPC reacts even during the first step, when the intermittent push is being applied. CTPC is therefore much stronger in rejecting even long-lasting intermittent pushes. } 
        \label{fig::intermittent}
\end{figure*}

In addition to analyzing self-stability, we characterize how well CTPC can recover intermittent pushes. In particular, we like to see how the timing of the external push will influence the state error. To this end, we consider the open-loop system, DLQR and CTPC controllers. The results are shown in Figure.\ref{fig::intermittent} over three consecutive steps. The external push is applied during the first step and error surfaces demonstrate the norm of error, calculated at touch down events. 

Figure.\ref{fig::intermittent} is clearly demonstrating that a push of same magnitude and duration might have more severe effect on the system if applied earlier in the phase. While the open-loop system is unstable, the other closed loop systems recover the push successfully, but with certain dynamics. The discrete DLQR controller produces quite the same error at the end of the first step, but due to delayed reaction, it overshoots in the second step. CTPC however uses internal capabilities of the robot to recover the push even during the first step. Such phenomena can be seen in Figure.\ref{fig::full_pushes} too. Comparing C and D footstep plans, CTPC almost requires only one step to recover the push while the aggressive DLQR requires two. This scenario perfectly shows that thanks to online reaction and proper configuration, CTPC outperforms discrete controllers.

\subsection{Controllable regions}

In terms of computation cost, due to online continuous-time control, CTPC requires more computation than the discrete DLQR. However, other controllers like MPC might require even more computations, due to multiple iterations required to optimize a certain cost function every time-step. The advantage of using such optimization-based controllers is mainly inclusion of inequality constraints. Without constraints, the controllable region for our linear yet unconstrained setup is in fact unlimited. Although input and state constraints are not considered, we would like to identify the set of states in which CTPC can operate safely. 

In this work, we only consider constraints on hip torque limits and footstep locations on the adult-size 3LP model. More constraints can be added on torque rates and velocities too. The torque limits are represented by simple boundaries ($\pm 80 [Nm]$) while next footstep locations should lie on a diamond region (with equal diameters of $1.7m$, compared to leg length of $0.9m$) centered at the stance foot location. The latter constraint could be more complex like using a circle, but here we use diamonds to preserve linearity. Note that we assume that foot cross-over is still possible and ignore self collisions. Considering realistic regions (like human) requires non-convex constraints. 

Due to the coupling of lateral and sagittal motions after adding diamond regions, we cannot split the 6-dimensional system into 3-dimensional subsystems. Therefore, it is not possible to completely visualize controllable regions. Additionally, these regions are now depending on the forward velocity as well. In other words, over faster speeds and thus longer stride lengths, possible perturbations are more limited compared to slower speeds. Therefore, we need to consider the effect of stepping frequency, speeds and dimensionality together.

Convex polyhedrons of controllable regions in this work are calculated for 10 consecutive steps which seem enough to give an approximate of the infinite shape. Going further increases the dimensionality and computation time while being less useful. Calculation of the full polyhedral is affordable only for 4-5 steps with overnight computations using a normal computer. Here however we use linear programming methods and a ray-casting technique to sample the convex hull around the projection of the full polyhedral on desired 2-dimensional subspaces. Here we limit the number of inter-phase samples for CTPC to 3 in order to make computations easier. We also calculate the maximal controllable region for any controller, using the same number of inter-phase samples. Note that if such maximal controllable region is calculated without inter-phase samples, it might become smaller than the region for CTPC.

\begin{figure}[]
        \centering
        \includegraphics[trim = 5mm 8.5mm 10mm 0mm, clip, width=0.5\textwidth]{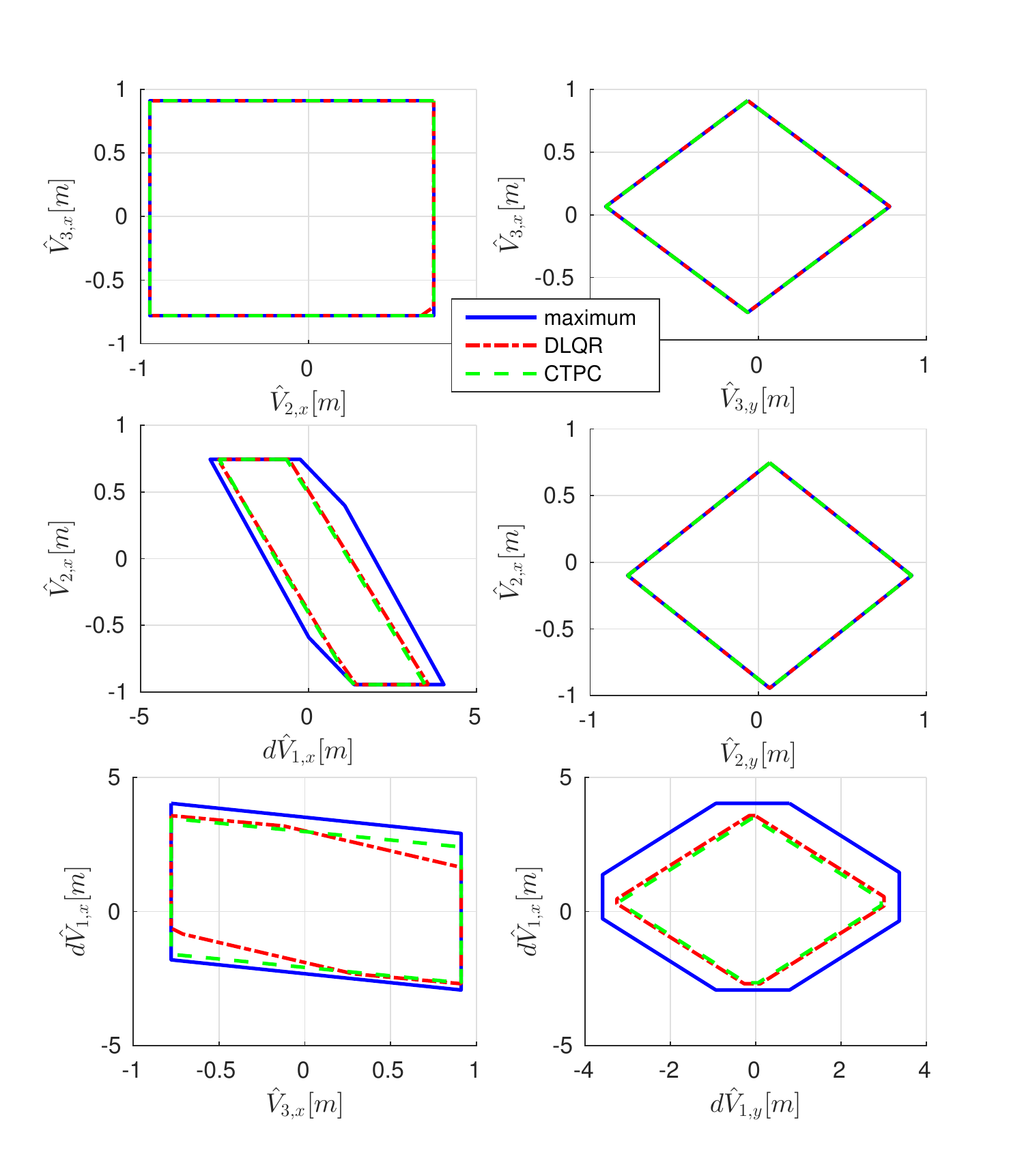}
        \caption{Different projections of the full 6-dimensional controllable regions calculated for the DLQR, CTPC and any other controller. Here we inspect maximum errors $\hat{V}_i$ of internal vectors $V_i$ demonstrated in Figure.\ref{fig::metric}. The reference gait is a pseudo-passive gait at $0.5 m/s$ and $3 step/s$. The DLQR and CTPC have quite similar controllable regions while the maximum any controller can do is slightly better in this specific walking conditions.} 
        \label{fig::sections}
\end{figure} 

Geometric constraints are active in faster speeds while torque limits are hit in faster frequencies. To demonstrate both constraints at the same time, we consider a less realistic case of walking at $0.5 m/s$ and $3 step/s$. It is impossible to view the full polyhedral from all perspectives, but the projection on important subspaces is shown in Figure.\ref{fig::sections}. Here, we take the representation of internal vectors as error measures (refer to Figure.\ref{fig::metric}) to give better intuition. The diamonds and torque boundary shapes are observable in some projections while others provide a less intuitive shape. One can observe that the DLQR and CTPC have quite similar controllable regions while the maximal region is a bit larger. 

\begin{figure}[]
        \centering
        \includegraphics[trim = 5mm 10mm 5mm 0mm, clip, width=0.5\textwidth]{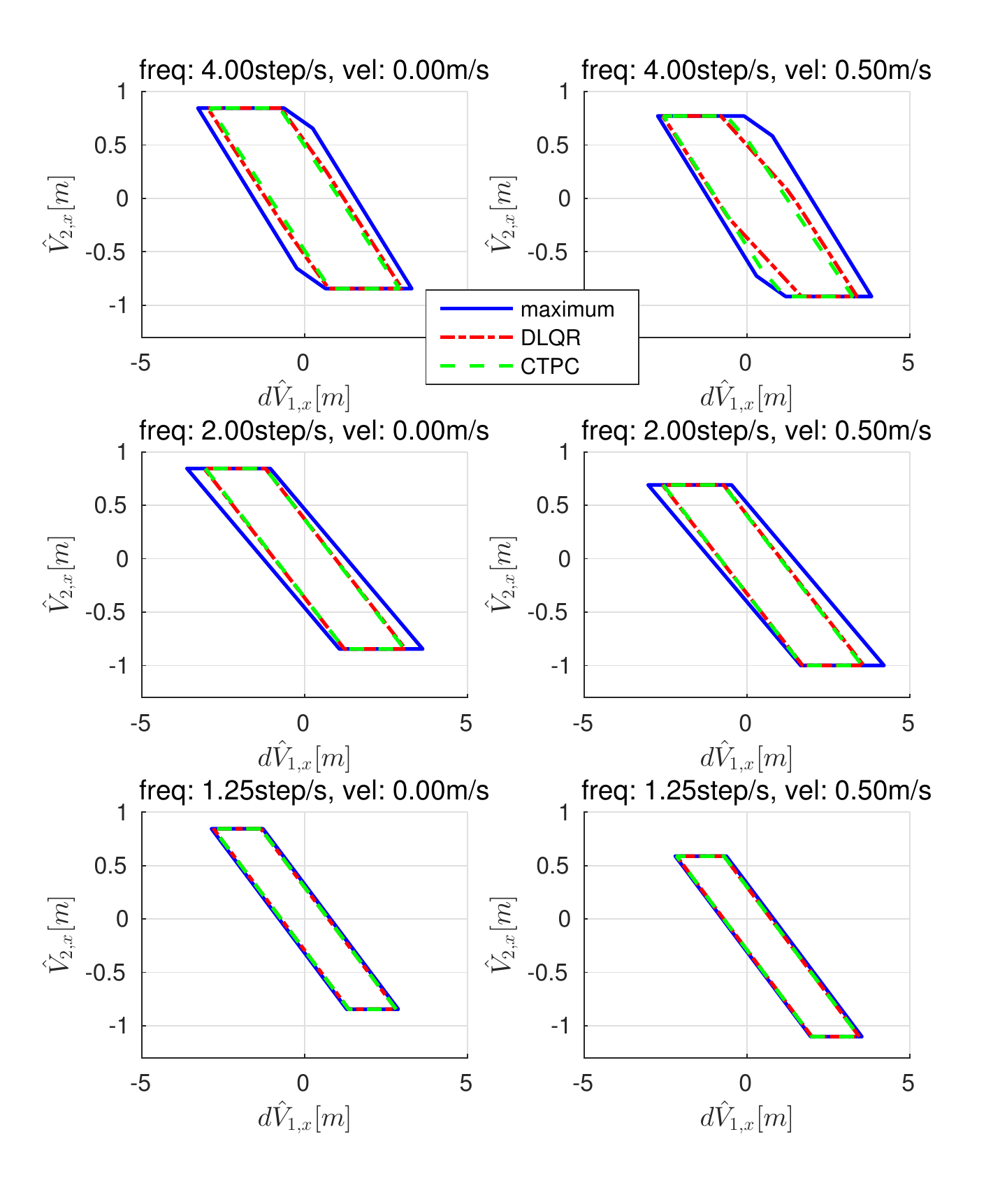}
        \caption{The effect of walking speed and frequency on controllable regions. This figure only demonstrates the projection of 6-dimensional polyhedrons on one subspace of position-velocity errors. As demonstrated over few choices of velocities and frequencies, the gait velocity mainly shifts the region while slower walking frequencies shrink it. } 
        \label{fig::stride_vel}
\end{figure}

Further, we are also interested to see the effect of walking speed and frequency on these regions. Again, it is hard to inspect the full-dimensional polyhedron, though we take a position/velocity projection which is more insightful. Figure.\ref{fig::stride_vel} demonstrates controllable regions for different choices of walking frequency and velocity. The velocity mainly shifts the controllable region while the frequency can shrink it due to stride-length limitations. Although the DLQR and CTPC can support most of the maximal controllable region, in higher frequencies they do not completely exploit the interplay between constraints and thus, they have more limited controllable regions. 

\subsection{Link to MPC framework}

Although our controller can easily handle intermittent pushes, we should acknowledge that still we can not introduce inequality constraints like torque limits and footstep length. These constraints can be used in a MPC framework where 3LP can be used as a core model to predict the future behavior. Constraints on modulating CoP and ankle torques are often used in literature during trajectory planning \cite{herdt2010online}. However hip torques are also important to be considered, specially for swing and torso dynamics. 3LP is the first linear template model that can provide information on hip-torques in an abstract level. 

Thanks to the analysis of controllable regions, we have now a more clear view of other possible controllers. We demonstrated that most of the time, CTPC can stabilize the system. Other controllers cover only slightly wider range of perturbed states. If the robot is severely perturbed, inequality constraints of controllable regions can simply determine if the perturbed state is recoverable or not. With such criteria, the algorithm could switch to other complicated controllers or even emergency cases. The analysis of controllable regions show that other controllers can rarely recover if CTPC fails. So most probably, the robot needs emergency actions. Therefore, MPC controllers cannot do much better than CTPC, if restricted to the aforementioned assumptions. However, exploring non-convex inequalities for footstep regions as well as timing adjustment could be possibly handled by non-linear and non-convex MPC controllers. 

\subsection{Different from intermittent control}

As we talk about intermittent disturbances, to avoid confusion, it is worth mentioning that the class of intermittent controllers \cite{gawthrop2007intermittent} are much different from our time-projecting controller. In this work, we propose an architecture based on per time-step observation and feedback paradigm. Inspired by certain experiments however, scientists propose a different control architecture for humans \cite{gawthrop2011intermittent}, successfully applied to simple systems too \cite{bhounsule2015discrete}. Instead of traditional observer-predictor-feedback paradigms, intermittent controllers use a kind of feedback which occasionally modifies certain parameters of the low level feed-forward or feedback controller. Such architecture can better deal with systems in which high level information is available with lower frequencies or larger delays, like humans. Compared to recent inverse dynamics methods \cite{faraji2014robust}, intermittent control is computationally less demanding of course, but versatility of this approach is yet questionable. Indeed, intermittent control in mixture with Neuro-Muscular model \cite{geyer2010muscle} might provide a good explanation of the control system in humans, specially over slow walking speeds. It is therefore a very interesting candidate for the control of 3LP in our future works.

\subsection{Exploiting swing dynamics}

In our previous works \cite{faraji2014robust} similarly, we calculated the evolution of current inter-phase error until the end of the stride and after, planned few steps using an MPC controller to stabilize the motion. The first footstep location found by MPC was then given to the lower level controllers to track. Therefore, only footstep adjustment was available as a control authority to capture the error. The remaining time of the phase was actually left without any action. In 3LP however, there is an ideal controller in the stance hip that keeps the torso always upright and of course influences the forward acceleration. Stance hip torques are influenced by swing hip torques and external disturbances too. Therefore, although the main role of swing hip torques is to adjust the footstep location, they can have a secondary influence on the forward acceleration as well. Hence in 3LP, swing dynamics can influence push recovery in two different ways, continuous (internal coupling) and discrete (footstep adjustment). The CTPC controller therefore is fully exploiting all capabilities of swing dynamics to recover pushes as fast as possible with minimal errors.

\section{Conclusion}

Motivated be the fact that in hierarchical controllers, the abstract plan should be matching the full dynamics as much as possible, we based our controller on 3LP model, introduced in \cite{faraji20163LP}. It provides many good features in terms of computation and more human-like gaits. In this paper, we take advantage of these features and propose a novel control architecture that works online to react against intermittent perturbations. Conventionally, other template models like inverted pendulum require time-integration, if used in predictive control paradigms. This is mainly due to non-linearity and lack of closed-form solutions. In 3LP however, linear equations provide transition matrices that can be used to find state evolution for any arbitrary period of time. In non-linear models, Poincar\'e maps \cite{poincare} linearize the model around a pre-calculated gait and provide a discrete control paradigm. The transition matrices in 3LP however let us easily increase the control update frequency and refine the control policy at every time-step. This new paradigm can handle intermittent pushes better, because it is less delayed. 

First, thanks to simple equations of discrete error dynamics, we setup a DLQR controller as core stabilizing expert. Then for each time-step in the middle of the continuous phase, we proposed an algorithm to project the error back in time and use the expertise of the DLQR controller. We optimized for the best time-projecting architectures and explored different features. Thanks to linear equations and exploitation of multiple alternative systems, we are able to decouple system dynamics from the effect of disturbances. Such decoupling with proper cost design let us minimize the influence of intermittent push timings. The discrete control paradigm however is blind to inter-phase disturbances and can observe their accumulation only at the end of the phase.  

Various analysis show that CTPC has similar stabilization and controllability properties with the DLQR controller. However it performs much better in recovery of intermittent pushes. Overall, advantages of our proposed architecture combined with the 3LP model are:
\begin{description*}
	\item[+] Resistant against intermittent perturbations.
	\item[+] No need for offline optimization. 
	\item[+] Computationally light compared to MPC.
	\item[+] Continuous-time policy refinement.
	\item[+] Better stabilization than the seeding DLQR controller.
	\item[+] Similar controllable region with DLQR.
	\item[+] Covering most of the maximal controllable region.
	\item[+] Speed-independent stability analysis. 
	\item[+] Optimal future behavior, thanks to the seeding DLQR.
	\item[+] Generic design for different model sizes.
\end{description*}
And disadvantages would be:
\begin{description*}
	\item[-] Fixed timing.
	\item[-] Lack of inequality-constraint support.
	\item[-] Requiring disturbance observer.
\end{description*}

Thanks to description of swing dynamics and other internal dynamic couplings, the proposed control paradigm provides clear failure analysis. In \cite{byl2008approximate} however since swing dynamics is absent, the hip controller tracks the desired attack angle only after mid-stance event. If the hip torque is applied earlier, it can cause the robot falling backward sometimes. 

CTPC can simply replace our previous MPC controller \cite{faraji2014robust}, bringing a lot of advantages. In future works, we want to use the proposed controller in combination with inverse dynamics and compare it with our previous MPC controllers. We would also like to exploit fast computational properties of the 3LP model to setup non-linear MPC controllers that can adjust the timing. Inclusion of non-convex constraints for avoiding self-collision is indeed another interesting future extension. Finally, in this paper we only explored external pushes and not other types of disturbances like uneven terrain for example. The formulation of 3LP is yet limited to impact-less locomotion which should be further improved to handle impacts and height variations as well. However, the concept of time-projection is yet applicable to other complex models, computationally challenging though if closed form solutions are not available. This paper is accompanied with a multimedia extension, demonstrating walking motions of speed tracking, discrete push and intermittent push recovery scenarios. All the codes used in this article as well as the multimedia extension are available online at \url{http://biorob.epfl.ch/page-99800-en.html}.

\section{Acknowledgments}
This work was funded by the WALK-MAN project (European Community's 7th Framework Programme: FP7-ICT 611832).

%\begin{thebibliography}{99}
%\bibitem[Kopka and Daly(2003)]{R1}
%Kopka~H and Daly~PW (2003) \textit{A Guide to \LaTeX}, 4th~edn.
%Addison-Wesley.
%
%\bibitem[Lamport(1994)]{R2}
%Lamport~L (1994) \textit{\LaTeX: a Document Preparation System},
%2nd~edn. Addison-Wesley.
%
%\bibitem[Mittelbach and Goossens(2004)]{R3}
%Mittelbach~F and Goossens~M (2004) \textit{The \LaTeX\ Companion},
%2nd~edn. Addison-Wesley.
%
%\end{thebibliography}

\bibliographystyle{IEEEtran}
\bibliography{Biblio}

\end{document}